%% file: iclr2026_conference.tex
\pdfoutput=1
\documentclass{article} % For LaTeX2e
\usepackage{iclr2026_conference,times}

\input{math_commands.tex}

\usepackage{hyperref}
\usepackage{url}
\usepackage{graphicx}
\usepackage{booktabs} % 需要在导言区引入此宏包
\usepackage{amsmath}  % 用于数学符号
\usepackage{geometry} % 调整页边距（可选，防止表格过宽超出边界）
\usepackage{enumitem}
\usepackage{caption}
\usepackage{adjustbox}
\usepackage{multirow}

\usepackage{rotating}  % 用于旋转表格
\usepackage{colortbl}  % 用于背景色
\usepackage[table]{xcolor} % 配合 colortbl

\makeatletter
\def\@evenhead{\hfil}
\def\@oddhead{\hfil}
\makeatother
\makeatletter
\newcommand{\customfootnotetext}[2]{%
  \begingroup
  \def\@thefnmark{#1}%
  \@footnotetext{#2}%
  \endgroup
}
\makeatother
\title{
    \vspace{-1cm} % 【省空间】把整个标题块往上提，吃掉顶部的无用白边
    \centering\Large An Industrial-Scale Insurance LLM Achieving Verifiable Domain Mastery and Hallucination Control without Competence Trade-offs
    \vspace{-0.2cm} % 【省空间】拉近标题和下方作者的距离
}

\author{Qian Zhu\thanks{Equal contribution} , 
  Xinnan Guo\footnotemark[1] , 
  Jingjing Huo\footnotemark[1] , 
  Jun Li\footnotemark[1] , 
  Pan Liu\footnotemark[1] ,
  \textbf{Wenyan Yang} ,
  \textbf{Wanqing Xu} ,
  \textbf{Xuan Lin}\thanks{Corresponding author. Xuan Lin led the overall technical direction of this project.} \\[0.25cm] 
  \multicolumn{1}{c}{\textbf{Ant Group}}
}

\iclrfinalcopy % Uncomment for camera-ready version, but NOT for submission.
\begin{document}

\maketitle
% --- 使用新命令插入邮箱，完美对齐且字体一致 ---
\customfootnotetext{$\ddagger$}{\{zq371417, daxuan.lx\}@antgroup.com}

\vspace{-0.2cm} 

\begin{abstract} 
Adapting Large Language Models (LLMs) to high-stakes vertical domains like insurance presents a significant challenge: scenarios demand strict adherence to complex regulations and business logic with zero tolerance for hallucinations. Existing approaches often suffer from a Competency Trade-off—sacrificing general intelligence for domain expertise—or rely heavily on RAG without intrinsic reasoning. To bridge this gap, we present \textbf{INS-S1}, an insurance-specific LLM family trained via a novel end-to-end alignment paradigm. Our approach features two methodological innovations: (1) A \textbf{Verifiable Data Synthesis System} that constructs hierarchical datasets for actuarial reasoning and compliance; and (2) A \textbf{Progressive SFT-RL Curriculum Framework} that integrates dynamic data annealing with a synergistic mix of Verified Reasoning (RLVR) and AI Feedback (RLAIF). By optimizing data ratios and reward signals, this framework enforces domain constraints while preventing catastrophic forgetting. Additionally, we release \textbf{INSEva}, the most comprehensive insurance benchmark to date (39k+ samples). Extensive experiments show that INS-S1 achieves \textbf{SOTA} performance on domain tasks, significantly outperforming DeepSeek-R1 and Gemini-2.5-Pro. Crucially, it maintains top-tier general capabilities and achieves a record-low 0.6\% hallucination rate (HHEM). Our results demonstrate that rigorous domain specialization can be achieved without compromising general intelligence. 
\end{abstract}

\section{Introduction}
\begin{figure}[t!]
    \centering
    \includegraphics[width=0.96\linewidth]{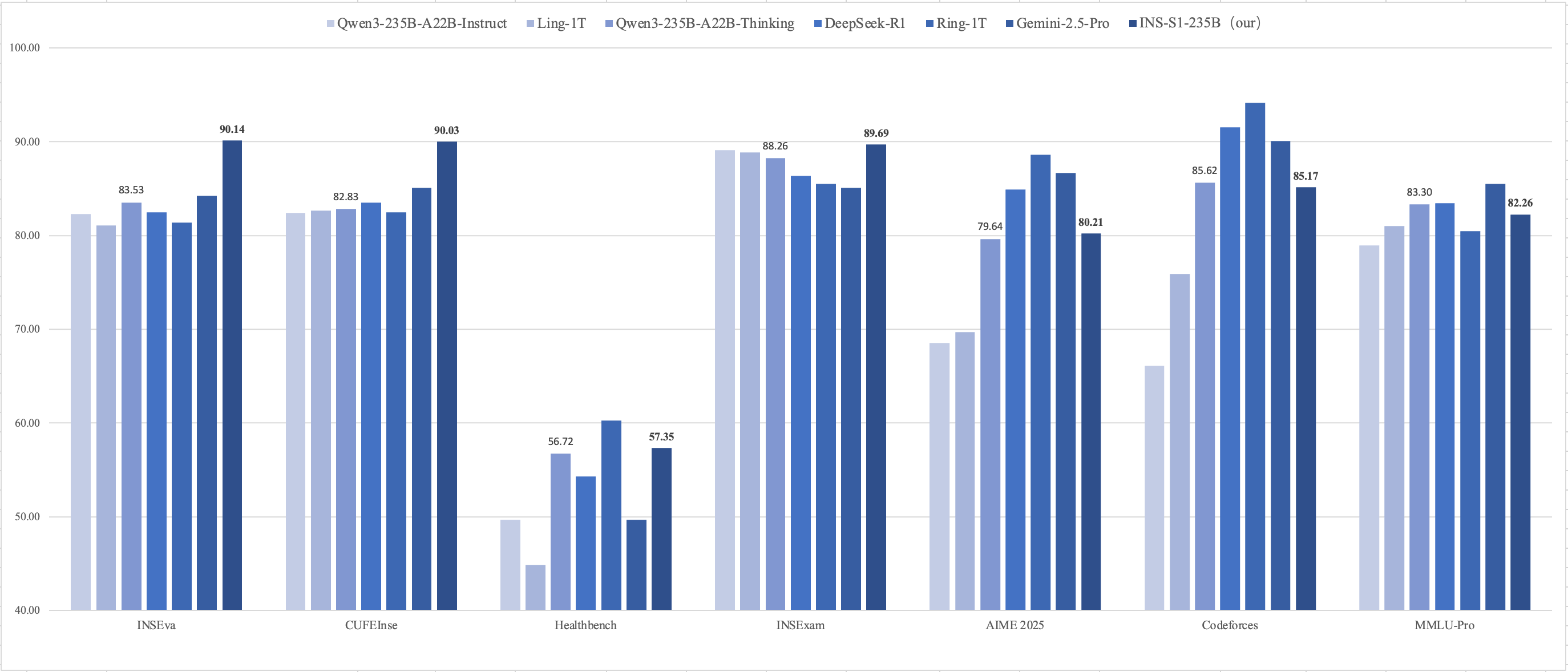}
    \caption{Performance comparison of INS-S1-235B against SOTA models.}
    \label{fig:performance_overview}
    \vspace{-18pt}
\end{figure}

Recent LLMs (e.g., DeepSeek-R1 \citep{deepseek2025deepseek}, Gemini \citep{comanici2025gemini}, Qwen3 \citep{yang2025qwen3}) have established new standards in reasoning. While domain models have advanced in finance and medicine, the \textbf{insurance sector remains underexplored}. Adapting to this domain demands strict compliance and a \textbf{robust mechanism} against hallucinations. Current solutions face three limitations:

\noindent\textbf{(1) Competency Trade-off and Shallow Adaptation:} 
Existing vertical models often sacrifice general intelligence for domain expertise. For instance, financial models like \textbf{DianJin-R1} \citep{zhu2025dianjin} suffer from severe \textbf{catastrophic forgetting} \citep{kirkpatrick2017overcoming}, struggling to balance specialized knowledge with general reasoning. Most adaptations rely on superficial RAG \citep{lewis2020retrieval} or naive fine-tuning, failing to internalize the multi-step reasoning required for actuarial pricing, resulting in models that lack deep reasoning capabilities.
\noindent\textbf{(2) High-Stakes Hallucination Risks:} Insurance decisions demand auditable reasoning paths. Current models often operate as ``black boxes,'' hallucinating policy terms or fictional cases. In a regulated industry, such errors lead not just to user distrust but to severe legal liabilities.
\noindent\textbf{(3) Lack of Standardized Evaluation:} The absence of a comprehensive, specialized benchmark hinders the iterative optimization of insurance LLMs. General financial benchmarks \citep{guo2025fineval} fail to capture the nuances of claim adjudication, underwriting logic, and regulatory compliance.

To address these challenges, we present \textbf{INS-S1}, a family of scalable insurance LLMs. Our approach integrates three core innovations:

\begin{itemize}[leftmargin=*]
    \setlength{\topsep}{0pt}
    \setlength{\itemsep}{0pt}
    \setlength{\parskip}{0pt}
    \setlength{\parsep}{0pt} 
    \item \textbf{Verifiable Insurance Data Synthesis System:} Transcending simple knowledge injection, we design a hierarchical dataset construction pipeline. It features fine-grained sub-task decomposition and a multi-source synthesis framework that enforces strict logic constraints, ensuring the model masters complex actuarial reasoning while maintaining general intelligence via self-distillation.
    
    \item \textbf{Progressive SFT-RL Curriculum Framework:} We propose a two-stage training paradigm that integrates Dynamic Data Annealing in SFT with a synergistic RL framework (RLVR \citep{lightman2023let} and RLAIF \citep{lee2023rlaif}). By optimizing data ratios and leveraging GRPO \citep{shao2024deepseekmath}/GSPO \citep{zheng2025group} optimization, this framework enables the model to master complex actuarial reasoning while strictly aligning outputs with insurance requirements.
    
    \item \textbf{INSEva Benchmark:} We release the industry's most comprehensive evaluation benchmark, comprising \textbf{39k} samples across 9 core dimensions and 50+ task types. Utilizing an automated ``LLM-as-a-Judge'' pipeline, INSEva provides a robust standard for assessing domain expertise.
\end{itemize}

Consequently, as shown in Figure~\ref{fig:performance_overview}, INS-S1-235B bridges the domain gap, scoring \textbf{90.14} on INSEva, surpassing general baselines like DeepSeek-R1 (82.47) by a significant margin. Crucially, it exhibits negligible degradation on general benchmarks compared to the backbone, while our RL alignment successfully constrains the hallucination rate to \textbf{0.6\%}. The detailed architecture is illustrated in Appendix~\ref{app:Architecture}.

\section{Data}

\subsection{Overview}
We implement a capability-driven data strategy centered on three pillars. \textbf{(1) Insurance Domain Mastery:} We construct compliant corpora using dual synthesis pipelines (No-Query and With-Query) to optimize domain task quality. \textbf{(2) Effective Hallucination Control:} We prioritize faithfulness via four atomic tasks and RAG adaptation, balancing rigor with expression. \textbf{(3) General Capability Preservation:} To prevent catastrophic forgetting, we curate a High-Fidelity Chinese dataset (spanning STEM, Math, Exams) and employ \textbf{Self-Distillation} to align general data with the base model's style.
\begin{figure}[t!]
    \centering
    \vspace{-10pt} % 压缩 Figure 上方的空白
    \includegraphics[width=0.95\linewidth]{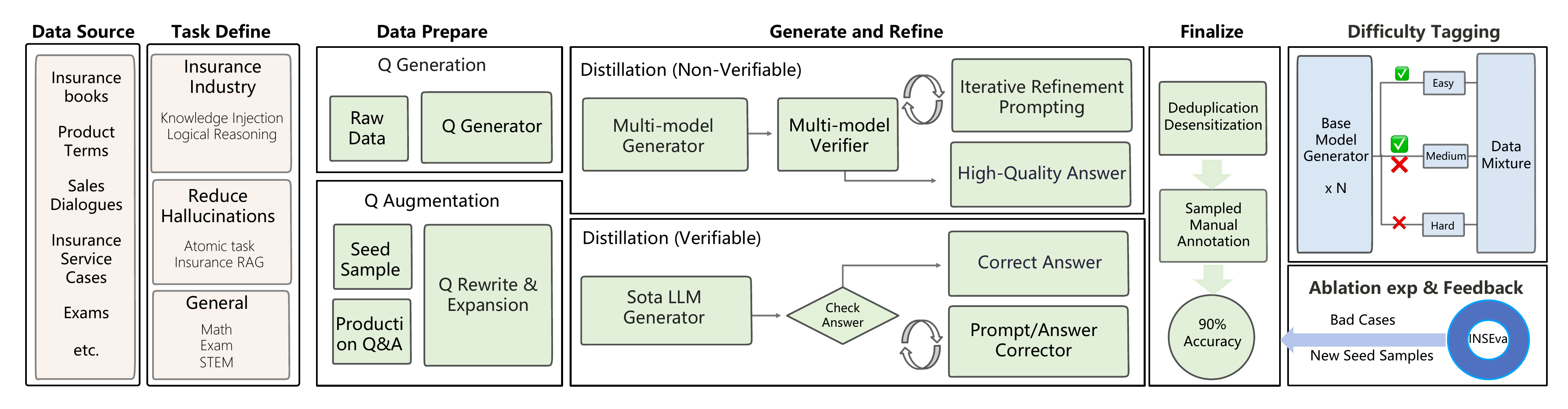}
    \caption{Data Synthesis Diagram.}
    \label{fig:placeholder}
    \vspace{-15pt} % 压缩 Figure 上方的空白
\end{figure}

\subsection{Data Construction Details}

\subsubsection{Insurance-Specific Data Construction}
\noindent\textbf{Knowledge Injection Tasks (No-Query).} We process multi-modal corpora (textbooks, regulations, tables) via a pipeline: Multi-modal Parsing $\to$ Markdown Representation $\to$ QA Generation. To prevent reasoning over-reliance for rote knowledge, we employ \textbf{Instruction Isolation} with ``Empty Think Tags'' (\texttt{<think></think>}), which improved accuracy on theory benchmarks by \textbf{3.11\%} compared to baselines.

\vspace{1pt} % 稍微手动加一点点极小的间距（比如2pt），防止太拥挤
\noindent\textbf{Cognitive Alignment Tasks (No-Query).} To address logical disconnects like boundary blurring, we propose \textbf{Business Logic Alignment}. In \textit{Query Generation}, inspired by Hard Negative Mining \citep{robinson2020contrastive}, we design \textbf{``Adversarial Distractors''}—options with subtle numerical or semantic alterations—to enforce precise discrimination. In \textit{Answer Generation}, we employ \textbf{SOP CoT} to enforce: (1) \textbf{Logic Explicitation}, decomposing judgment into a verification pipeline (Entity $\to$ Attribute $\to$ Compliance); and (2) \textbf{Correction-based Reasoning}, mandating explicit clause citations for self-verification.

\begin{figure}[htbp]
    \centering
    \vspace{-10pt} % 压缩 Figure 上方的空白
    \begin{minipage}{0.48\textwidth} % 稍微加大宽度利用率 (0.46 -> 0.48)
        \centering
        \includegraphics[width=\linewidth]{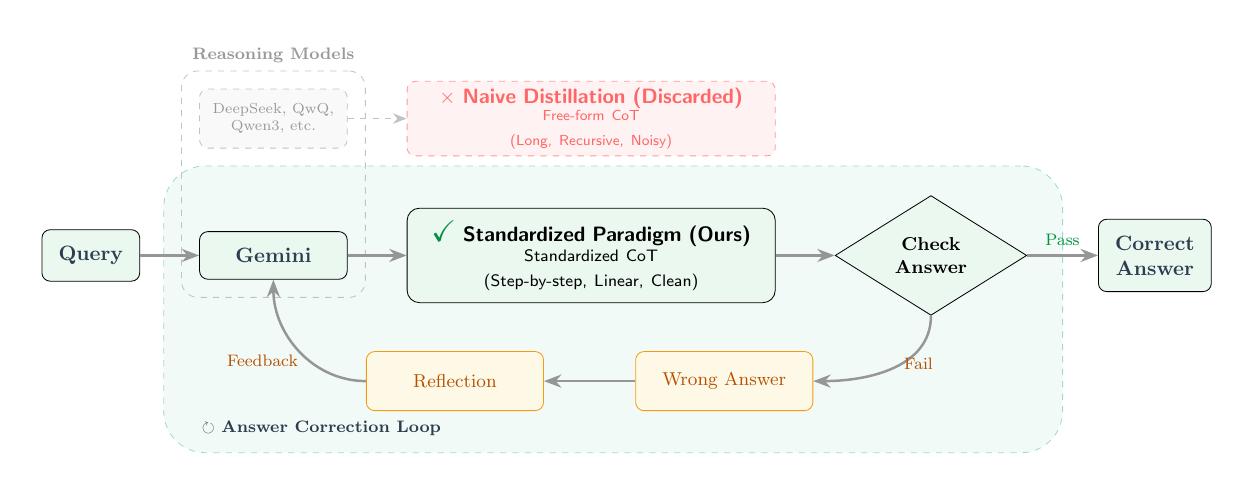}
        \caption{Answer Correction Loop.} % 缩短 caption
        \label{fig:answer_loop}
    \end{minipage}
    \hfill
    \begin{minipage}{0.48\textwidth} % 稍微加大宽度利用率 (0.53 -> 0.48)
        \centering
        \includegraphics[width=\linewidth]{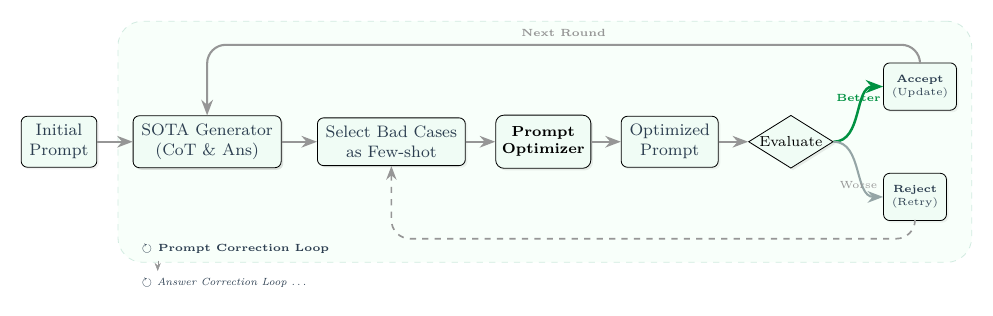}
        \caption{Prompt Correction Loop.} % 缩短 caption
        \label{fig:prompt_loop}
    \end{minipage}
    \vspace{-6pt} % 压缩 Figure 上方的空白
\end{figure}

\noindent\textbf{Complex Reasoning Tasks (With-Query).} We target two domains. For \textbf{Actuarial Math}, we mitigate noisy free-form CoT (Fig.~\ref{fig:answer_loop}) via: (1) \textbf{Standardized Reasoning}, aligning derivations with linear steps (similar to PoT) \citep{chen2022program}, yielding a \textbf{+10.8\%} gain; and (2) an \textbf{Answer Correction Loop} (Self-Refine) \citep{madaan2023self} that iteratively executes a \textit{Generate-Verify-Reflect-Rewrite} pipeline. For \textbf{Underwriting \& Claims}, we handle implicit constraints via: (1) an APE-inspired \textbf{Prompt Correction Loop} \citep{zhou2022large}, refining prompts based on validation errors (Fig.~\ref{fig:prompt_loop}) to boost accuracy ($80\!\to\!85$); and (2) \textbf{SOP-CoT Injection}, enforcing a structured logic path (Info Extraction $\to$ Risk ID $\to$ Rule Detection $\to$ Conclusion) to ensure long-tail robustness.

\subsubsection{Hallucination Mitigation Data}
We adopt a phased approach, migrating the model to the insurance domain via low-hallucination QA and refusal samples. Ultimately, we optimize using GRPO with a Faithfulness reward model, mitigating reward hacking by incorporating real scenario queries.

\noindent\textbf{Atomic Capability Enhancement.} We decompose RAG into four tasks: (1) \textbf{Knowledge Boundary ID} using CAIL2019 \citep{duan2019cjrc} to train refusal when evidence is lacking; (2) \textbf{Knowledge Selection} via DRCD \citep{shao2018drcd} with distractors for precise location; (3) \textbf{Summarization} combining CNN/DailyMail \citep{see2017get} and CASSum with \textbf{RLAIF} for factual distillation; and (4) \textbf{Self-Check} leveraging CHEF \citep{hu2022chef} for post-generation hallucination correction.

\begin{figure}[htbp]
    \centering
    \vspace{-20pt} % 压缩图片上方留白
    \begin{minipage}{0.48\textwidth}
        \centering
        \includegraphics[width=\linewidth]{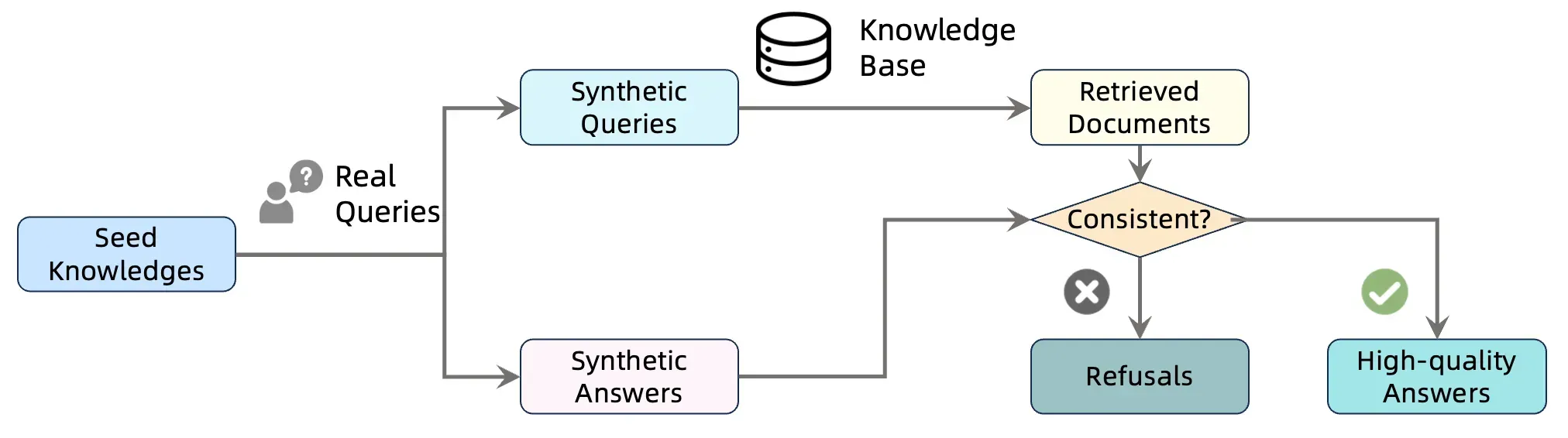}
        \caption{RAG Task Adaptation.}
        \label{fig:rag_task} % 修正了 Label
    \end{minipage}
    \hfill
    \begin{minipage}{0.48\textwidth}
        \centering
        \includegraphics[width=\linewidth]{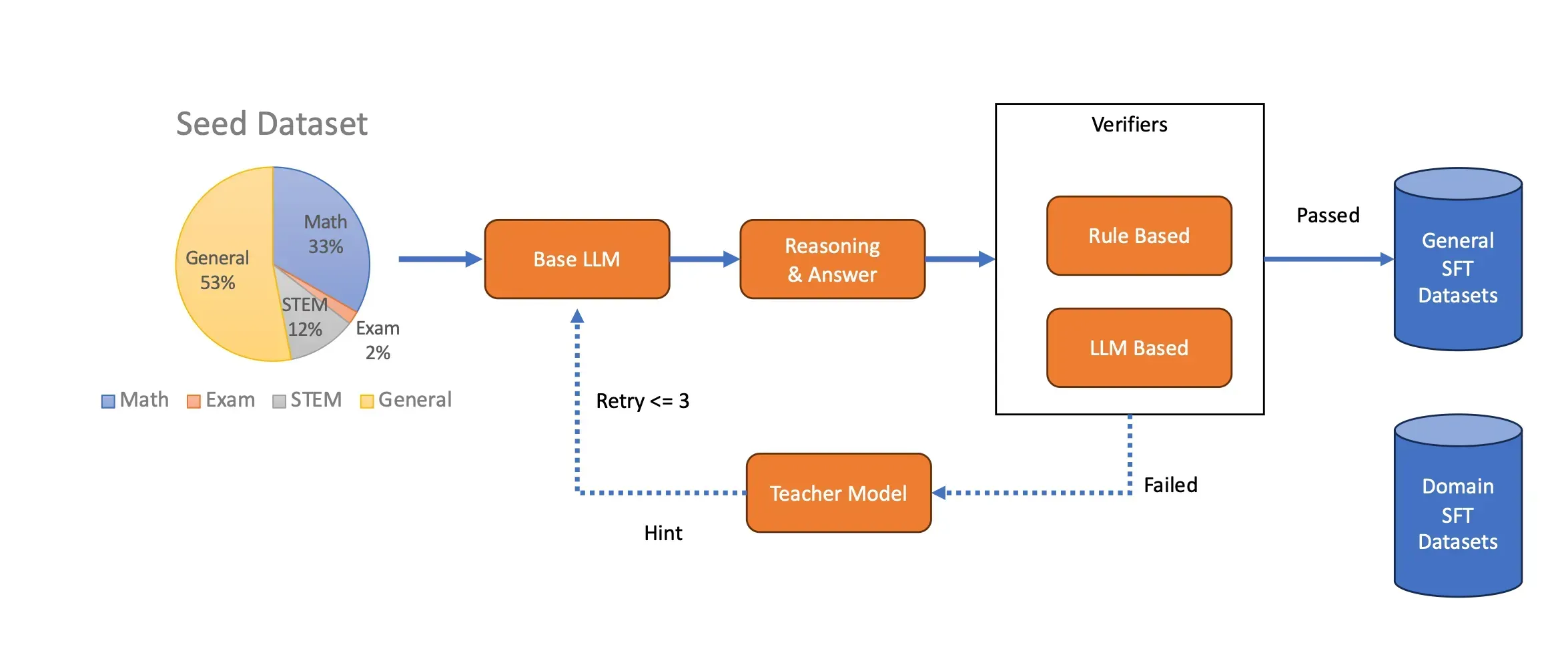}
        \caption{General Data Composition.}
        \label{fig:general_data} % 修正了 Label
    \end{minipage}
    \vspace{-10pt} % 压缩图片下方留白
\end{figure}

\vspace{1pt} % 微调间距
\noindent\textbf{RAG Task Adaptation.} We synthesize QA pairs from FAQs, using an LLM validator to route consistent samples to generation training and inconsistent ones to refusal datasets (Fig.~\ref{fig:rag_task}). This strategy improved factuality by \textbf{18.21\%} over baselines.

\noindent\textbf{Response Style Optimization.} To mitigate reward hacking (e.g., rote copying) during RL, we integrate expression signals into the reward model (see Sec.\ref{sec:expression_optimization}). We sample real single- and multi-turn insurance dialogues to ensure the model balances accuracy with interaction quality.

\subsubsection{General Data}
To mitigate \textbf{Catastrophic Forgetting}, we deploy a \textbf{Self-Distilled} \citep{yang2024self} dataset (110k samples) as a ``Capability Stabilizer,'' constructed via: (1) \textbf{High-Fidelity Alignment} ensuring distribution consistency via base model generation; (2) \textbf{Multi-Stage Verification} combining programmatic checks (Math/Code) with human sampling; (3) \textbf{Complexity Screening} prioritizing medium-length instructions to avoid noisy long CoTs; and (4) \textbf{Closed-Loop Monitoring} to purge samples dynamically upon detecting benchmark decay.
\section{Training}
\begin{figure}[htbp]
    \centering
    \vspace{-10pt} % 压缩 Figure 下方的空白
    \includegraphics[width=0.7\linewidth]{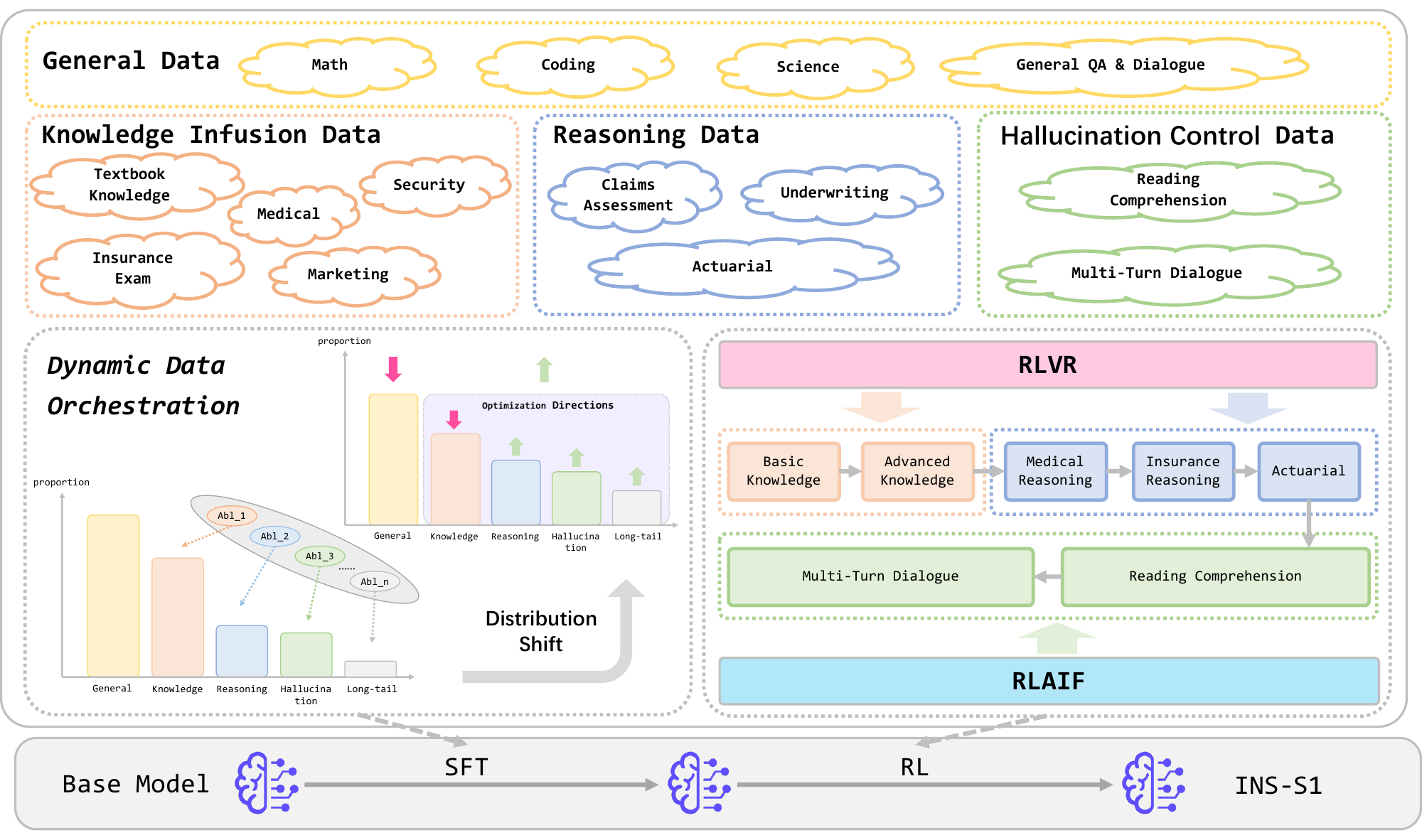}
    \caption{SFT-RL Training Framework}
    \label{fig:SFT-RL Training}
    \vspace{-10pt} % 压缩 Figure 下方的空白
\end{figure}

We construct \textbf{INS-S1} via a progressive SFT-RL framework rooted in Curriculum Learning (Fig.~\ref{fig:SFT-RL Training}).

\textbf{SFT Stage (Breadth to Depth):} We employ dynamic data orchestration. The initial phase mixes fundamental and complex tasks, while the later phase introduces Data Annealing \citep{zhou2023lima}, significantly up-sampling high-difficulty reasoning. This shifts the model's focus from general patterns to conquering core domain challenges.
\textbf{RL Stage (Dual Alignment):} We optimize via two mechanisms: (1) RLVR (Verified Reasoning), utilizing Process Supervision to verify step-wise correctness in complex scenarios (e.g., actuarial science); and (2) RLAIF (AI Feedback), leveraging strong model feedback to reward Faithfulness, ensuring output reliability and curbing hallucinations.

\subsection{SFT}
\subsubsection{Trainable Data Ratios in SFT}
We propose a dynamic mixing strategy guided by \textbf{Component-wise Ablation Feedback}, utilizing convergence rates as proxies for difficulty. We down-sample rapidly converging tasks and up-sample complex reasoning. As validated through iterations (v1 vs. v2) shown in Table~\ref{tab:ablation_training}, increasing insurance logical reasoning data from 7.5\% to 11\% significantly improved domain performance while maintaining general balance \citep{xie2023doremi}. The final data distribution of SFT data is shown in Appendix~\ref{app:Data_Distribution}.

\subsubsection{Curriculum Learning via Dynamic Data Orchestration}
\label{sec:annealing_strategy} % <--- 加上这一句 Label
We decouple SFT into two stages to balance stability and plasticity.
\textbf{Stage 1: Foundation \& Stability.} We mix Self-distilled general data with domain data under a constant learning rate to prevent catastrophic forgetting.
\textbf{Stage 2: Domain-Adaptive Annealing.} We introduce a Dynamic Re-weighting strategy \citep{hu2024minicpm} to reconstruct data distribution: (1) General vs. Domain: Reducing general data to 25\% while increasing domain data to $>30\%$; (2) Difficulty Stratification: Up-sampling complex reasoning and hallucination mitigation tasks to 50\% over simple tasks (30\%); (3) Long-tail Boosting \citep{liu2019large}: Aggressively up-sampling sparse sub-capabilities ($<1$k samples) to fit low-frequency knowledge.

\subsubsection{Challenges in Fitting Long CoT Data}
\label{sec:long_cot_challenges}
Fitting long Chain of Thought (CoT) poses challenges for smaller models. We find that CoT distilled from Gemini—with superior density—outperforms native Qwen3-32B CoT in enhancing Qwen3-14B. Furthermore, while we evaluated Dynamic Fine-Tuning (DFT) \citep{wu2025generalization} and Critic Fine-Tuning (CFT) \citep{wang2025critique}, they yielded lower gains than standard SFT. Consequently, we adopt ``High Knowledge Density CoT combined with Standard SFT'' as our optimal strategy.

\begin{table}[htbp]
\centering
\caption{Ablation study on training strategies.}
\label{tab:ablation_training}

% 1. 缩小字体和行高
\small
\setlength{\tabcolsep}{3pt} % 极限压缩列间距
\renewcommand{\arraystretch}{0.8}
% 2. 强制适配宽度
\begin{tabular}{l | c ccccccccc}
\toprule % 5. 关键修改：加回最上方的横线
\textbf{Model Variant} & \textbf{Avg.} & IDK & IMI & IUC & ILR & ITU & IPE & ISC & IMG & ISD \\
\midrule
% 核心实验组
v2-Stage1 & 84.01 & 85.99 & 80.52 & 89.56 & \textbf{75.69} & 84.19 & \textbf{84.14} & 88.67 & \textbf{84.14} & \textbf{85.54} \\
v1-Stage1 & 81.86 & \textbf{86.47} & 80.26 & 89.58 & 72.00 & 84.72 & 83.22 & 89.56 & 81.76 & 80.14 \\
\midrule
% 对比组 (Stage 2)
v1-Stage2 (\textit{Curriculum}) & \textbf{84.39} & 85.99 & \textbf{84.19} & \textbf{90.29} & 75.23 & \textbf{86.52} & 83.63 & \textbf{92.40} & 83.48 & 83.97 \\
v1-Stage2 (\textit{Mixing}) & 83.01 & 85.02 & 80.27 & 89.82 & 74.69 & 84.60 & 83.20 & 89.22 & 82.08 & 82.97 \\
\midrule
% 基线
Qwen3-32B (Baseline) & 78.31 & 84.06 & 80.24 & 83.64 & 70.87 & 81.91 & 84.18 & 69.93 & 83.74 & 75.93 \\
\bottomrule
\end{tabular}
\vspace{-10pt} % 压缩 Figure 下方的空白

\end{table}

\subsection{RL}

\subsubsection{Task-Sequential Curriculum Learning}
Unlike random mixing, we implement \textbf{Hierarchical Curriculum Learning} \citep{soviany2022curriculum} in the RL phase, following ``Macro-Stage Progression and Micro-Difficulty Ascension.'' Training is divided into three tiers: (1) \textit{Knowledge Consolidation}; (2) \textit{Complex Reasoning} (domain-interleaving); and (3) \textit{Alignment \& Robustness} (faithfulness) \citep{ji2023survey}. Comparisons confirm that this ordered knowledge injection significantly outperforms random mixing in logical reasoning and instruction following (Tab.~\ref{tab:ablation_training}).

\subsubsection{Synergistic Use of RLVR and RLAIF}
\label{sec:expression_optimization}
To address diverse tasks, we construct a \textbf{Hybrid Reward Modeling Framework} splitting training into parallel streams:

\noindent\textbf{1. RLVR (Verifiable Tasks):} For tasks with ground truth (e.g., actuarial formulas), we employ dual verification: (1) \textbf{Rule-based Matching} (Regex) for unique solutions; and (2) \textbf{Semantic Equivalence}, utilizing Qwen3-30B in \textbf{Direct Inference Mode} to rapidly determine mathematical/logical equivalence.

\noindent\textbf{2. RLAIF (Open Tasks):} For open-ended scenarios, we use an \textbf{RLAIF} framework with Qwen3-30B in \textbf{CoT Mode} to evaluate: (1) \textbf{Factuality:} Prioritizing faithfulness to RAG context and penalizing mechanical copying; (2) \textbf{Professionalism:} Ensuring logic coherence and safety guardrails; (3) \textbf{Expression:} Scoring conciseness to avoid a "robotic tone." More details in Appendix~\ref{app:Use of RLVR and RLAIF}.

\subsubsection{RL Algorithm Selection and Modification}
We adopt an \textbf{Architecture-aware} post-training strategy.
\textbf{(1) Algorithm Selection:} For the Dense Model (32B), we use GRPO to leverage intra-group advantages. For the MoE Model (235B), we adopt GSPO to mitigate routing instability. We exclude DAPO \citep{yu2025dapo} due to its computational overhead on unranked data.
\textbf{(2) Efficiency-Aware Length Reward:} To enforce the \textbf{Principle of Parsimony}—penalizing verbose CoT that adds no value—we design a conditional reward $R = R_{acc} \cdot (\alpha + \beta \cdot R_{L})$, where $R_L = \text{clip}(\frac{L_{max} - L(x_i)}{L_{max} - L_{min}}, 0, 1)$ serves as a piecewise linear truncation function effectively only when the answer is correct.

More details of ablation experiments are provided in the Appendix~\ref{app:ablation_study}

\section{Experiments}
\subsection{Benchmark}
\begin{figure}[htbp]
    \centering
    \vspace{-10pt} % 压缩 Figure 下方的空白
    \includegraphics[width=0.7\linewidth]{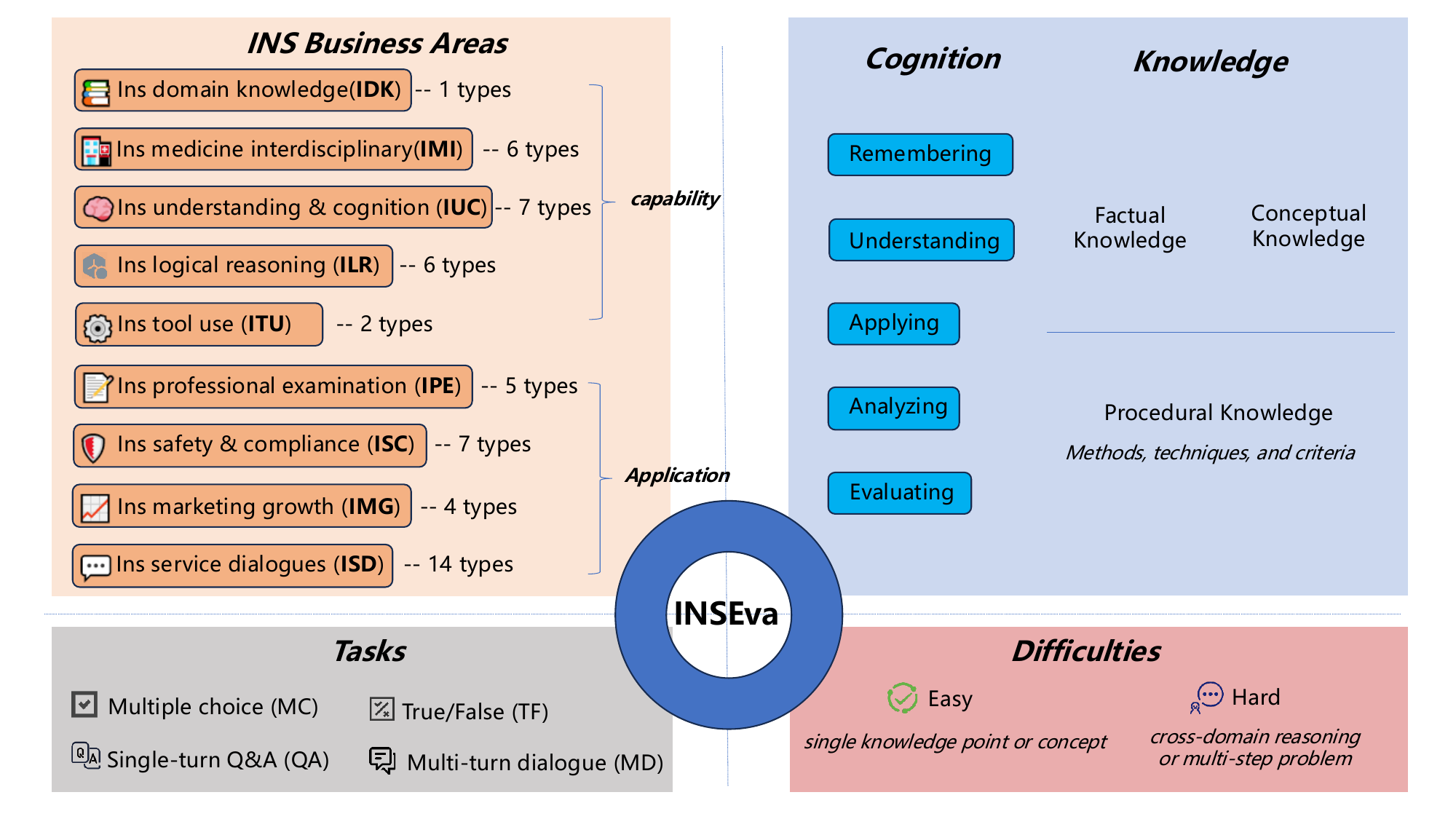}
    \caption{The taxonomy of INSEva benchmark.}
    \label{fig:placeholder}
    \vspace{-10pt} % 压缩 Figure 下方的空白
\end{figure}
We introduce \textbf{INSEva}, a comprehensive benchmark comprising 39,166 high-quality examples from authoritative sources (e.g., exams, regulations) and stratified by a multi-dimensional taxonomy (Business, Format, Difficulty, Cognition). Within this framework, we adopt task-specific metrics: \textbf{Accuracy} for deterministic items and \textbf{Faithfulness} (via an LLM-based verification pipeline) for open-ended generation.Appendix~\ref{app:INSEva_details}

To ensure robustness, we also evaluate on \textbf{CUFEInse} \citep{cufeinse2024} (a complementary academic benchmark), the hallucination benchmark \textbf{HHEM} \citep{vectara2024hhem}, and 40 general benchmarks (e.g., \textbf{AIME25} \citep{mathai2025aime25}, \textbf{Codeforces} \citep{quan2025codeelo}, \textbf{MMLU-Pro} \citep{wang2024mmlu}), assessing whether domain specialization impacts general reasoning (Appendix~\ref{app:Generation_Benchmark_Details}).

\subsection{Baseline}

We compare models across four categories: (1) \textbf{General models without explicit reasoning:} Qwen3-Instruct (30B/235B), Kimi-K2, Ling-1T \citep{team2025ling}, and GPT-4o \citep{hurst2024gpt} serve as broad generalization baselines. (2) \textbf{General models with reasoning capabilities:} Reasoning-enhanced models including Qwen3-Thinking (32B/80B/235B), DeepSeek-R1, Ring-1T \citep{team2025ring}, and Gemini-2.5-Pro are used to assess logical performance. (3) \textbf{Financial models with reasoning abilities:} Dianjin-R1 serve as domain adaptation baselines. (4) \textbf{Insurance models with reasoning capabilities:} Our proposed \textbf{INS-S1 (32B/235B)} features a hybrid architecture trained on extensive insurance and general instructions.

\subsection{Training Settings}
\begin{table}[htbp]
\centering
\vspace{-10pt}
\caption{Hyperparameters and compute resources for SFT and RL training stages.}
\label{tab:training_params}

% 1. 极小字体 (7pt)
\small
% 2. 极扁行高
\renewcommand{\arraystretch}{0.75} 
% 3. 紧凑列间距
\setlength{\tabcolsep}{2.5pt} 

\begin{tabular}{l c | c c c}
\toprule 
\textbf{Model} & Stage & GPU Resources & Learning Rate & Epochs \\
\midrule
\multirow{2}{*}{INS-S1-32B} & SFT & $96 \times \text{H20-3e}$ & 2e-6 & 2 const. + 0.3 anneal. \\
& RL & $96 \times \text{H20-3e}$ & 1e-6 & 1 \\
\midrule
\multirow{2}{*}{INS-S1-235B} & SFT & $512 \times \text{910C}$ & 5e-6 & 2 const. + 0.3 anneal. \\
& RL & $160 \times \text{H20-3e}$ & 1e-6 & 1 \\
\bottomrule 
\end{tabular}
\vspace{-15pt}
\end{table}

\subsection{Main Results}

\begin{table*}[t!]
\caption{Main results on insurance and general benchmarks.}
\centering
% \scalebox{0.8}{ % 适配页面宽度，可按需微调
\adjustbox{max width=\textwidth}{ % 自动适配页面宽度
\begin{tabular}{l c | c c c c | c | c c c c  | c | c}
% 列格式：l(左对齐) c(居中) |(竖线)，Finova-Avg./GPQA-Avg.间添加竖线
\toprule
\textbf{Model} & Params & \multicolumn{5}{c|}{Insurance} & \multicolumn{5}{c|}{General} & Two Avg. \\
& & \textbf{INSEva} & CUFEInse & Healthbench & INSExam & Avg. & AIME25 & Codeforces & MMLU-Pro & etc.. & Avg. \\
\midrule
\multicolumn{10}{c}{\textit{General Models (No Reasoning)}} \\ % 分类标题居中
\midrule % 替换cmidrule，确保横线与表格同宽
% \addlinespace[2pt]
Qwen3-30B-A3B-Instruct & 30B & 79.59& 79.63& 49.26& 84.08& 73.14& 62.29& 61.15& 73.58& -& 76.40 & 74.77\\
Qwen3-235B-A22B-Instruct & 235B & 82.32& 82.43& 49.70& 89.09& 75.89& 68.54& 66.10& 78.95& -& 79.91	& 77.90 \\
Kimi-K2-Instruct & 1TB & 81.78 &82.27 &48.54 &87.96 &75.14 &48.28 &55.03 &81.27 &- &78.17 & 76.65 \\
Ling-1T & 1TB & 81.09 &82.67 &44.84 &88.87 &74.37 &69.69 &75.89 &81.02 &- &80.70	 & 77.53 \\
GPT-4o & -- & 79.41 & 79.72 & 37.73 & 79.05 & 68.98 & 8.80 & 23.39 & 63.82 &- 	& 67.78	 & 68.38 \\
\midrule
\multicolumn{10}{c}{\textit{General Models (With Reasoning)}} \\ % 分类标题居中
\midrule % 等宽横线
\addlinespace[2pt]
Qwen3-32B (think) & 32B & 80.88 &81.18 &47.96 &83.42 &73.36 &71.67 &80.91 &78.38 &- &79.22 & 76.29 \\
Qwen3-Next-80B-A3B-Thinking & 80B & 73.07 &75.13 &52.79 &86.41 &71.85 &85.47 &84.77 &81.40 &- &80.79 & 76.32 \\
Qwen3-235B-A22B-Thinking & 235B & 83.53 &82.83 &56.72 &88.26 &77.84 &79.64 &85.62 &83.30 &- &81.40 &79.62 \\
DeepSeek-R1 & 671B & 82.47 &83.53 &54.33 &86.40 &76.68 &84.90 &91.52 &83.48 &- &82.17 & 79.43 \\
Ring-1T & 1TB & 81.40 &82.45 &\textbf{60.27} &85.50 &77.41 &\textbf{88.65} &\textbf{94.14} &80.47 &- &\textbf{82.34} &79.01 \\
Gemini-2.5-Pro & -- & 84.23& 85.09& 49.69& 85.10& 76.03& 86.70& 90.11& \textbf{85.54}& - & 81.98	& 79.01\\
\midrule
\multicolumn{10}{c}{\textit{Financial Models (With Reasoning)}} \\ % 分类标题居中
\midrule % 等宽横线
\addlinespace[2pt]
Dianjin-R1-32B & 32B & -  & 80.00	& - & - & - & 22.60 & 40.06 & 75.78 &  - & 68.82 &  68.82 \\
\midrule
\multicolumn{10}{c}{\textit{Insurance Models (With Reasoning)}} \\ % 分类标题居中
\midrule % 等宽横线
\addlinespace[2pt]
INS-S1-32B & 32B & 88.42 & 88.5 & 47.43 & 88.60 & 78.24 & 76.41 & 77.62 & 77.73 & - & 80.08	& 79.16 \\
INS-S1-235B & 235B & \textbf{90.14} &\textbf{90.03} &57.35 &\textbf{89.69} &\textbf{81.80} &80.21 &85.17 &82.26 &- &81.77 &\textbf{81.79} \\
\bottomrule
\end{tabular}}
\label{tab:Ins_Results}
\end{table*}

Results validate the effectiveness of our Verifiable Data Synthesis and Curriculum Learning paradigms, \textbf{demonstrating scalability across model sizes} and establishing a new SOTA (Tab.~\ref{tab:Ins_Results},~\ref{tab:INSEva_Results}).

\noindent\textbf{1. Domain Proficiency \& Parameter Efficiency.} INS-S1-235B ranks first on INSEva with a score of \textbf{90.14}, significantly outperforming DeepSeek-R1 (82.47) and Gemini-2.5-Pro (84.23). Notably, our medium-scale INS-S1-32B (88.42) surpasses the comparably sized Qwen3-32B (80.88) and even larger models (Ling-1T), proving that high-quality instruction sets enable medium models to overcome the ``competence dilution'' of general LLMs. INS-S1 leads in 7 out of 8 sub-dimensions (e.g., IMI, IUC, ISD), exceeding baselines by over \textbf{5 points}.

\vspace{2pt}
\noindent\textbf{2. Improvements in Reasoning \& Safety.} We address critical gaps in general models: (1) \textbf{Actuarial Reasoning (ILR):} While general models struggle with quantitative rules (Qwen3-80B-A3B: 54.25), INS-S1-32B achieves \textbf{81.27} (235B: 83.61) via our specialized Actuarial Math data. (2) \textbf{Safety \& Dialogue (ISC/ISD):} Leveraging hallucination mitigation data and RLVR/RLAIF, INS-S1-235B scores \textbf{91.39} in compliance and \textbf{92.15} in dialogue, far surpassing DeepSeek-R1 (80.10) and Qwen3-235B (81.40).

\vspace{2pt}
\noindent\textbf{3. Positive Transfer to General Capabilities.} Contrary to the ``zero-sum'' trade-off, INS-S1 retains top-tier general reasoning (AIME25: 80.21, Codeforces: 85.17). INS-S1-32B even outperforms its backbone on general benchmarks (80.08 vs. 79.22). This suggests a deep positive transfer where rigorous insurance logic enhances general mathematical/logical reasoning. Notably, we also achieved a SOTA score of \textbf{0.6\%} on the HHEM hallucination benchmark (Appendix ~\ref{app:HHEM}).

\begin{table*}[htbp]
\caption{Main results for LLMs on INSEva across different Business Areas.}
\centering
\small % 1. 使用小号字体
\renewcommand{\arraystretch}{0.8} % 2. 压缩行高 (默认是1.0，0.85会明显变扁)
% \scalebox{0.8}{ % 适配页面宽度，可按需微调
\adjustbox{max width=\textwidth}{ % 自动适配页面宽度
\begin{tabular}{l c | c c c c c c c c c | c}
% 列格式：l(左对齐) c(居中) |(竖线)，Finova-Avg./GPQA-Avg.间添加竖线
\toprule
\textbf{Model} & Params & \multicolumn{9}{c}{Business Areas} & INSEva. \\
& & IDK & IMI & IUC & ILR & ITU & IPE & ISC & IMG & ISD \\
\midrule
\multicolumn{11}{c}{\textit{General Models (No Reasoning)}} \\ % 分类标题居中
\midrule % 替换cmidrule，确保横线与表格同宽
% \addlinespace[2pt]
Qwen3-30B-A3B-Instruct & 30B & 81.64 & 75.50 & 83.96 & 66.25 & 81.38 & 80.30 & 72.86 & 87.82 & 83.79 & 79.59 \\
Qwen3-235B-A22B-Instruct & 235B & 86.76 & 79.87 & 	84.24 & 72.43 & 85.33 & 86.22 & 74.26 & 88.83 & 84.95 & 82.32 \\
Kimi-K2-Instruct & 1TB & 83.57 & 81.45 & 86.41 & 71.61 & 88.18 & 85.67 & 77.14 & 90.55 & 80.35 & 81.78 \\
Ling-1T & 1TB & 86.47 & 77.54 & 82.02 & 74.41 & 82.85 & 87.02 & 72.19 & 87.39 & 83.00 & 81.09\\
GPT-4o & -- & 79.71 & 75.47 & 83.83 & 68.90 & 82.52 & 73.19 & 71.96 & 83.72 & 86.04 & 79.41 \\
\midrule
\multicolumn{11}{c}{\textit{General Models (With Reasoning)}} \\ % 分类标题居中
\midrule % 等宽横线
\addlinespace[2pt]
Qwen3-32B (think) & 32B & 82.13 & 80.23 & 84.06 & 79.19 & 81.91 & 84.39 & 69.54 & 84.10 & 81.13 &  80.88 \\
Qwen3-Next-80B-A3B-Thinking & 80B & 86.47 & 76.50 & 81.32 & 54.25 & 76.43 & 85.55 & 77.89 & 76.54 & 67.27 & 73.07 \\
Qwen3-235B-A22B-Thinking & 235B & 85.99 & 82.68 & 82.20 & 80.55 & \textbf{89.32} & 88.98 & 79.98 & \textbf{92.31} & 81.40  & 83.53\\
DeepSeek-R1 & 671B & 82.13 & 81.47 & 83.20 & 80.77 & 89.02 & 86.55 & 80.10 & 87.08 & 80.26 & 82.47 \\
Ring-1T & 1TB & 83.09 & 81.51 & 82.64 & 78.07 & 84.72 & 87.16 & 78.38 & 91.57 & 77.48 & 81.40 \\
Gemini-2.5-Pro & -- & 83.09 & 82.81 & 86.16 & 74.06 & 87.77	 & 88.58 & 76.97 & 91.77 & 86.16 & 84.23  \\
\midrule
\multicolumn{11}{c}{\textit{Financial Models (With Reasoning)}} \\ % 分类标题居中
\midrule % 等宽横线
\addlinespace[2pt]
Dianjin-R1-32B & 32B & 85.02 & 75.19 & 83.65 & 39.40 & 87.65 & 83.46 & 74.29 & 80.90 & 81.30 & 75.71 \\
\midrule
\multicolumn{11}{c}{\textit{Insurance Models (With Reasoning)}} \\ % 分类标题居中
\midrule % 等宽横线
\addlinespace[2pt]
INS-S1-32B & 32B & 86.96 & 89.57 & 90.85 & 81.27 & 89.02 & 88.51 & 89.47 & 88.41 & 89.46 & 88.42 \\
INS-S1-235B & 235B & \textbf{87.92} & \textbf{90.16} & \textbf{91.81} & \textbf{83.61} & 85.55 &\textbf{90.26} &\textbf{91.39} & 91.43 &\textbf{92.15} & \textbf{90.14} \\
\bottomrule
\end{tabular}}
\label{tab:INSEva_Results}
\end{table*}

\section{Conclusion and Future Work}

We presented \textbf{INS-S1}, demonstrating that \textbf{Verifiable Data Synthesis} and \textbf{Progressive SFT-RL Curriculum} achieve verifiable domain mastery and hallucination control without sacrificing general capabilities, validating its efficacy in large-scale industrial applications (Appendix~\ref{app:Deployment}). However, a gap persists in processing non-standardized actuarial reasoning, highlighting the distinction between general math and domain rigor. Future work focuses on three directions: (1) \textbf{Robust Long-Context Instruction:} Addressing ``Lost-in-the-Middle'' issues in ultra-long windows ($>128$k) to enable precise ``Light Training, Heavy Prompting''; (2) \textbf{Scalable Subjective Rewards:} Developing \textbf{RLAIF}-based ``Human-Feeling'' reward models to align with human values with minimal annotation; and (3) \textbf{Agentic Capabilities:} Enhancing \textbf{Function Calling} to bridge the gap from consultation to closed-loop business transactions via API integration.

\bibliography{iclr2026_conference}
\bibliographystyle{iclr2026_conference}

\appendix

\section{Ablation Study}
\label{app:ablation_study}

\subsection{Knowledge Injection Tasks}
\begin{table}[htbp]
    \centering
    \caption{Ablation Study on Knowledge Injection Strategies. Evaluated on the CUFE-Inse (Insurance Theory) benchmark.}
    \label{tab:knowledge_ablation}
    \begin{adjustbox}{max width=0.8\linewidth} % 限制宽度，防止表格过宽
    \begin{tabular}{l c c}
    \toprule
    \textbf{Method} & \textbf{Accuracy (\%)} & \textbf{$\Delta$ vs Baseline} \\
    \midrule
    Qwen3-14B (Baseline) & 77.82 & - \\
    + Distilled CoT (Q\{T\}A) & 78.70 & +0.88 \\
    + Template-Constrained CoT (Q\{DT\}A) & 77.66 & -0.16 \\
    \textbf{+ Empty Think Tag (Ours)} & \textbf{80.93} & \textbf{+3.11} \\
    \bottomrule
    \end{tabular}
    \end{adjustbox}
\end{table}

\noindent\textbf{Analysis of Knowledge Injection Strategies:}
To optimize domain knowledge absorption, we explored three formatting paradigms. Results (Table~\ref{tab:knowledge_ablation}) indicate:
\begin{itemize}[leftmargin=*]
    \item \textbf{Superiority of Instruction Isolation:} The \textbf{Empty Think Tag} strategy (\texttt{<think></think>}) achieved the highest gain (\textbf{+3.11\%}). This confirms that for rote knowledge tasks, forcing explicit reasoning paths is unnecessary and potentially distracting; simple memorization yields better results.
    \item \textbf{Inefficacy of Forced Reasoning:} Introducing \textbf{Distilled CoT} provided only marginal gains (+0.88\%), while rigid \textbf{Template Constraints} (Q\{DT\}A) actually degraded performance (-0.16\%). This suggests that forcing a base model to mimic external reasoning patterns for atomic knowledge disrupts its native representation space.
    \item \textbf{Conclusion:} Consequently, we adopt Instruction Isolation with Empty Think Tags as the standard protocol for knowledge injection.
\end{itemize}

\subsection{Cognitive Alignment Tasks}

\begin{table}[htbp]
    \centering
    \caption{Ablation Study on Cognitive Alignment Tasks. Incorporating SOP CoT and Business-Trap Distractors yields significant gains in domain understanding.}
    \label{tab:alignment_ablation}
    \begin{adjustbox}{max width=0.85\linewidth}
    \begin{tabular}{l c c}
        \toprule
        \textbf{Model Configuration} & \textbf{Liability Analysis} & \textbf{Slot ID} \\
        \midrule
        Qwen3-8B (Base) & 62.70 & 85.64 \\
        + \textbf{SFT (SOP CoT + Traps)} & \textbf{92.25} & \textbf{97.52} \\
        \bottomrule
    \end{tabular}
    \end{adjustbox}
\end{table}

\noindent\textbf{Cognitive Alignment Strategy Analysis.} To bridge the gap between intuitive judgment and strict business logic, we employed a dual-stage optimization:
\begin{itemize}[leftmargin=*]
    \item \textbf{Query Generation (Hard Negatives):} Inspired by \citet{robinson2020contrastive}, we synthesized \textbf{``Business-Trap Distractors''}---options with subtle tampering (e.g., ``Cash Value'' vs. ``Sum Assured'')---to enforce precise concept discrimination.
    \item \textbf{Answer Generation (SOP CoT):} We introduced \textbf{Logic Explicitation} via a multi-step verification pipeline (Entity $\to$ Attribute $\to$ Compliance) and \textbf{Correction-based Reasoning}, mandating explicit clause citations for self-verification.
\end{itemize}
\noindent\textbf{Result:} As shown in Table~\ref{tab:alignment_ablation}, these strategies achieved a massive \textbf{+29.55} point gain ($62.70 \to 92.25$) in Insurance Liability Analysis, validating the efficacy of structured logic injection.

\subsection{Complex Reasoning Tasks}
\paragraph{Actuarial Math}
Actuarial tasks demand rigorous, multi-step derivation. We observed that naive distillation yields ``Free-form CoT'' containing significant cognitive noise (redundant exploration), which hinders SFT convergence. To address this, we propose a \textbf{Standardized Reasoning Paradigm} aligned with Program-of-Thought (PoT) \citep{chen2022program}, enforcing linear, step-by-step derivations.
\textbf{Effectiveness:} As shown in Table~\ref{tab:actuarial_cot}, under identical data scales (6k), our standardized approach achieves a \textbf{+10.8\%} gain over Free-form CoT on InsEva 1.0, confirming the necessity of structured reasoning.

\begin{table}[htbp]
    \centering
    \caption{Ablation Study: Standardized vs. Free-form CoT (InsEva 1.0).}
    \label{tab:actuarial_cot}
    \begin{adjustbox}{max width=0.85\linewidth}
    \begin{tabular}{l c c}
        \toprule
        \textbf{Model Configuration} & \textbf{Ins. Actuary} & \textbf{FinMath (CAA)} \\
        \midrule
        Qwen3-14B (Base) & 46.73 & 55.68 \\
        + 6k (Free-form CoT) & 60.23 & 62.62 \\
        \textbf{+ 6k (Standardized CoT)} & \textbf{71.03} & \textbf{71.59} \\
        \bottomrule
    \end{tabular}
    \end{adjustbox}
\end{table}

For intractable queries, we implement a ``Generate-Verify-Reflect-Rewrite'' loop (Fig.~\ref{fig:answer_loop}) inspired by Self-Refine \citep{madaan2023self}. As detailed in Table~\ref{tab:hard_case_scaling}, scaling this high-quality corrected data ($1.2\text{w} \to 1.7\text{w}$) yields continuous marginal gains across complex tasks like Actuarial Practice.

\begin{table}[htbp]
    \centering
    \caption{Impact of Scaling Corrected Hard Case Data (InsEva 2.0).}
    \label{tab:hard_case_scaling}
    \begin{adjustbox}{max width=0.95\linewidth}
    \begin{tabular}{l c c c}
        \toprule
        \textbf{Model Configuration} & \textbf{Ins. Actuary} & \textbf{CAA Exam} & \textbf{Actuarial Practice} \\
        \midrule
        Qwen3-14B (Base) & 52.34 & 71.00 & 37.82 \\
        + 1.2w SFT & 71.96 & 74.75 & 47.90 \\
        \textbf{+ 1.7w SFT} & \textbf{74.77} & \textbf{76.00} & \textbf{52.10} \\
        \bottomrule
    \end{tabular}
    \end{adjustbox}
\end{table}

\paragraph{Underwriting \& Claims Reasoning}
Unlike calculation-centric actuarial tasks, underwriting and claims demand strict \textbf{Rule Adherence} within long contexts. We address the challenge of ``implicit rules''---expert intuition missing from explicit clauses---via a \textbf{Prompt Correction Loop} (see Fig.~\ref{fig:prompt_loop}). Inspired by APE \citep{zhou2022large} and Automatic Instruction Optimization \citep{pryzant2023automatic}, this mechanism iteratively diagnoses error cases to explicate hidden business logic. 

As shown in Table~\ref{tab:prompt_stats}, three iterations boosted test accuracy significantly (e.g., Underwriting: $67\% \to 93\%$). Furthermore, we enforce \textbf{SOP-CoT} (\textit{Info Extraction $\to$ Risk ID $\to$ Rule Detection $\to$ Conclusion}), which, combined with the refined data, yields substantial performance gains over the baseline (Table~\ref{tab:underwriting_ablation}).

\begin{table}[htbp]
    \centering
    \caption{Effectiveness of Prompt Correction Loop on Data Production. Iterative refinement significantly boosts Test Set accuracy and ensures high yield rates for the Training Set.}
    \label{tab:prompt_stats}
    \begin{adjustbox}{max width=0.9\linewidth}
    \begin{tabular}{l c c}
        \toprule
        \textbf{Production Stage} & \textbf{Claims Data} & \textbf{Underwriting Data} \\
        \midrule
        \textit{Test Set Accuracy} & & \\
        \quad Initial Prompt & 80.0\% & 67.0\% \\
        \quad \textbf{Prompt Correct Loop ($\times 3$)} & \textbf{85.0\%} & \textbf{93.0\%} \\
        \midrule
        \textit{Train Set Quality} & & \\
        \quad Initial Prompt (Acc.) & 80.0\% & 60.7\% \\
        \quad Final Prompt (Acc.) & 86.1\% & 81.0\% \\
        \quad \textbf{Answer Correct Loop (Yield)} & \textbf{98.6\%} & \textbf{92.5\%} \\
        \bottomrule
    \end{tabular}
    \end{adjustbox}
\end{table}

\begin{table}[htbp]
    \centering
    \caption{Ablation Study on Underwriting \& Claims Reasoning. SOP-CoT combined with Prompt Correction yields significant improvements.}
    \label{tab:underwriting_ablation}
    \begin{adjustbox}{max width=0.85\linewidth}
    \begin{tabular}{l c c}
        \toprule
        \textbf{Model Configuration} & \textbf{Claims Reasoning} & \textbf{Underwriting Reasoning} \\
        \midrule
        Qwen3-14B (Base) & 70.50 & 67.62 \\
        \textbf{+ 1.5w SFT (w/ SOP)} & \textbf{86.00} & \textbf{73.81} \\
        \bottomrule
    \end{tabular}
    \end{adjustbox}
\end{table}

\subsection{RAG Task Adaptation}
The specific process (see Fig.~\ref{fig:rag_task}), and the experimental results are supplemented (Table~\ref{tab:rag_factuality_ablation}).
\begin{table}[htbp]
  \centering
  \caption{Ablation Study on Factuality Improvements. Combining Atomic Task Data with specific Insurance RAG Data yields the highest performance gain.}
  \label{tab:rag_factuality_ablation}
  \begin{tabular}{l c}
    \toprule
    \textbf{Model Configuration} & \textbf{Factuality Score} \\
    \midrule
    Qwen3-14B (Base) & 74.80 \\
    \quad + Atomic Task Data & 81.34 ($\uparrow$8.74\%) \\
    \quad + Insurance RAG Data & 84.99 ($\uparrow$13.62\%) \\
    \textbf{\quad + Both (Full SFT)} & \textbf{88.42 ($\uparrow$18.21\%)} \\
    \bottomrule
  \end{tabular}
\end{table}

\subsection{Data Annealing Strategy}
The specific process is detailed in Section~\ref{sec:annealing_strategy}, and the experimental results are supplemented in Table~\ref{tab:annealing_ablation}.
\begin{table}[htbp]
  \centering
  \caption{Ablation Study of Data Annealing Strategy. Strategic downsampling during the annealing phase yields the highest performance gains compared to constant sampling or uniform downsampling.}
  \label{tab:annealing_ablation}
  \begin{adjustbox}{max width=\linewidth}
  \begin{tabular}{l c c c c c c}
    \toprule
    \textbf{Model Configuration} & \textbf{Overall} & \textbf{Ins. Theory} & \textbf{Ind. Und.} & \textbf{Agent App.} & \textbf{Safety} & \textbf{Rigor} \\
    \midrule
    INS-S1-32B (Constant) & 88.20 & 84.16 & \textbf{91.79} & 85.76 & 86.34 & 95.65 \\
    INS-S1-32B (Anneal: Uniform) & 88.11 ($\downarrow$0.09) & 84.31 & 91.30 & 86.02 & 86.45 & 95.72 \\
    \textbf{INS-S1-32B (Anneal: Strategic)} & \textbf{88.63 ($\uparrow$0.43)} & \textbf{85.24} & 91.52 & \textbf{87.48} & \textbf{86.48} & \textbf{96.30} \\
    \bottomrule
  \end{tabular}
  \end{adjustbox}
\end{table}

\subsection{Long Cot Data}
To address the optimization difficulties associated with fitting long Chain-of-Thought (CoT) data (as discussed in Section~\ref{sec:long_cot_challenges}), we evaluated distinct fine-tuning strategies. The comparative results are presented in Table~\ref{tab:reasoning_performance}.

\begin{table}[htbp]
  \centering
  \caption{Performance Comparison of Different Fine-tuning Strategies on Actuarial Reasoning Tasks.}
  \label{tab:reasoning_performance}
  % lccc: Left aligned first column, Centered remaining three
  \begin{tabular}{l c c c}
    \toprule
    \textbf{Model} & \textbf{CAA Exam} & \textbf{Actuarial MCQ} & \textbf{Ins. Actuary} \\
    \midrule
    Qwen3-14B (Base) & 71.00 & 37.82 & 52.34 \\
    Qwen3-14B-DFT & 63.00 & 29.41 & 61.68 \\
    Qwen3-14B-CFT & \textbf{73.75} & 41.60 & 59.81 \\
    Qwen3-14B-SFT & 73.50 & \textbf{44.96} & \textbf{71.03} \\
    \bottomrule
  \end{tabular}
\end{table}

\subsection{Scaling Inference-Time Compute}
To evaluate INS-S1's thinking capability on complex reasoning tasks, we analyze the impact of the RL inference phase Rollout parameter ($N$). Table~\ref{tab:rollout_ablation} compares the backbone model (Qwen3-32B) with INS-S1-32B under different Rollout configurations ($N=4$ and $N=16$).

\begin{table}[htbp]
\centering
\caption{Impact of Inference-Time Compute (Rollout $N$) on Actuarial and Reasoning Tasks. Performance is reported on specific sub-tasks of INSEva.}
\label{tab:rollout_ablation}
\begin{adjustbox}{max width=\linewidth}
\begin{tabular}{l c c c c c c}
\toprule
\textbf{Model} & \textbf{Overall} & \textbf{Actuarial} & \textbf{CAA Exam} & \textbf{Ins. Accountant} & \textbf{Claims Reas.} & \textbf{Undwrt. Reas.} \\
\midrule
Qwen3-32B (backbone) & 81.18 & 53.78 & 75.25 & 59.81 & 71.50 & 66.67 \\
INS-S1-32B ($N=4$) & 89.39 & 65.13 & 79.75 & 69.16 & 76.50 & 67.62 \\
INS-S1-32B ($N=16$) & \textbf{89.77} & \textbf{65.97} & \textbf{85.50} & \textbf{78.50} & \textbf{77.50} & \textbf{69.05} \\
\bottomrule
\end{tabular}
\end{adjustbox}
\end{table}

\noindent\textbf{Analysis:} Results indicate a significant positive correlation between inference-time compute and domain reasoning capability. Specifically, increasing $N$ from 4 to 16 yielded substantial gains in highly logical tasks (e.g., Chinese Actuary Exam \textbf{+5.75\%} and Insurance Accountant Exam \textbf{+9.34\%}). This suggests that ``Scaling Test-time Compute'' is a viable and effective strategy for complex domain tasks requiring multi-step deduction.

\subsection{Negative Constraint-based Reward Shaping}
\label{app:HHEM}
\textbf{Metric Definition: HHEM.} To strictly quantify and suppress hallucinations (crucial in insurance), we adopt the \textbf{Hughes Hallucination Evaluation Model (HHEM)} \citep{vectara2024hhem} as the primary evaluation metric and a proxy for reward signals. HHEM calculates a ``hallucination rate'' (lower is better) by assessing whether the generated summary is fully entailed by the source document.

\begin{table}[htbp]
\centering
\caption{Trade-off Analysis: HHEM Hallucination Rate vs. Domain \& General Capabilities. INS-S1 achieves a record-low hallucination rate (0.6\%) with negligible degradation on general benchmarks.}
\label{tab:general_capabilities}
\begin{adjustbox}{max width=\linewidth}
\begin{tabular}{l c c c c c c c c c}
\toprule
\multirow{3}{*}{\textbf{Model}} & \textbf{Safety} & \textbf{Domain} & \multicolumn{7}{c}{\textbf{General Capabilities}} \\
\cmidrule(lr){2-2} \cmidrule(lr){3-3} \cmidrule(lr){4-10}
 & \textbf{HHEM} & \textbf{INSEva} & \textbf{Avg} & \multicolumn{2}{c}{\textbf{Math}} & \textbf{Code} & \textbf{Instr.} & \multicolumn{2}{c}{\textbf{Lang. Und.}} \\
 & ($\downarrow$) & ($\uparrow$) & & \scriptsize{AIME24} & \scriptsize{AIME25} & \scriptsize{CodeForces} & \scriptsize{IFEval} & \scriptsize{MMLU} & \scriptsize{GPQA} \\
\midrule
Qwen3-32B & 2.8\% & 78.67 & \textbf{79.25} & 80.73 & 71.67 & \textbf{80.91} & \textbf{86.27} & \textbf{89.29} & \textbf{66.64} \\
INS-S1-32B (+HHEM) & \textbf{0.6\%} & \textbf{87.42} & 78.70 & \textbf{80.94} & \textbf{71.72} & 79.13 & 86.10 & 88.88 & 65.44 \\
\bottomrule
\end{tabular}
\end{adjustbox}
\end{table}

\noindent\textbf{Method: Iterative Reward Shaping.} Balancing factuality (low HHEM score) and response quality in RLHF is a core challenge. Following the ``Principle of Parsimony,'' we implemented a negative constraint-based reward shaping strategy. Instead of complex positive rewards, we iteratively introduced penalty terms to curb ``Reward Hacking'' behaviors observed during training:

\begin{itemize}[leftmargin=*]
    \item \textbf{V1 (Hallucination Mitigation):} Basic reward derived from HHEM scores to penalize unsupported content. $\to$ \textit{Issue: Model started copying prompt text verbatim.}
    \item \textbf{V2 (+Anti-duplication):} Penalty for excessive n-gram overlap. $\to$ \textit{Issue: Code-switching (Mixed Chinese/English) appeared to bypass overlap checks.}
    \item \textbf{V3 (+Language Consistency):} Constraint on language uniformity. $\to$ \textit{Issue: Repetitive structural patterns.}
    \item \textbf{V4 (+Diversity \& Naturalness):} Final penalty for repetitive patterns to ensure natural expression.
\end{itemize}

\subsection{Real-world Deployment and Effectiveness}
\label{app:Deployment}
\begin{figure}[htbp]
    \centering
    \vspace{-10pt}
    \includegraphics[width=1\linewidth]{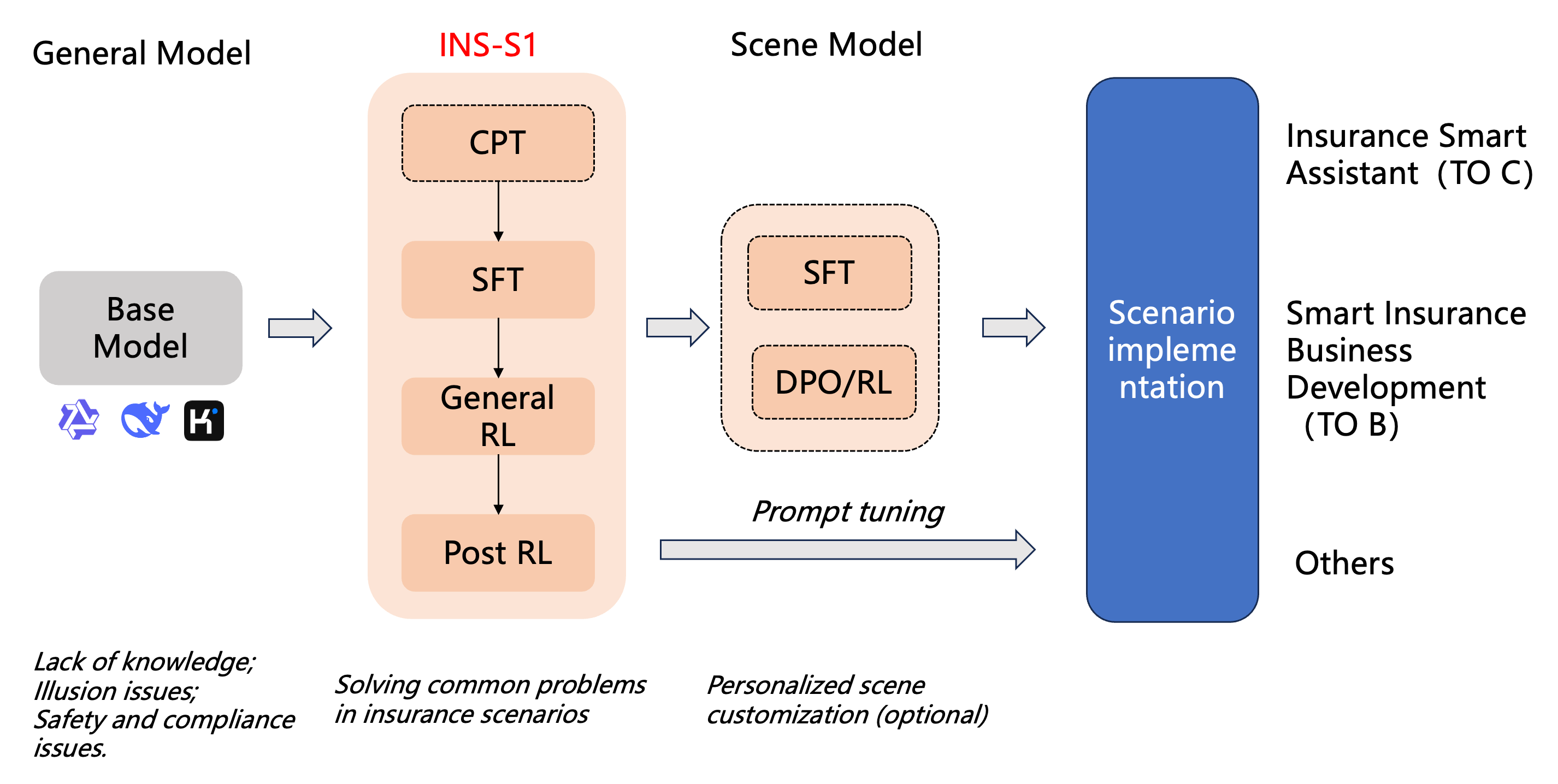}
    \vspace{-10pt}
    \caption{Real-world Deployment Workflow.}
    \label{fig:deployment}
    \vspace{-10pt}
\end{figure}

Finally, we evaluated the practical value of INS-S1 in downstream business scenarios. As shown in Figure~\ref{fig:deployment}, INS-S1 acts as the core ``Scenario Model,'' solving generic domain challenges before being adapted to specific \textit{Insurance Intelligent Assistant (To C)} tasks via lightweight RL or Prompt fine-tuning.

We conducted a strict double-blind \textbf{Side-by-Side (SBS)} evaluation using real business queries. Metrics included Professional Rate, Factuality Rate, and \textbf{Human Preference Win-Tie Rate}.

\begin{table}[htbp]
\centering
\caption{Side-by-Side (SBS) Evaluation in Real-world Insurance Smart Assistant Scenarios.}
\label{tab:scenario_performance}
\begin{adjustbox}{max width=\linewidth}
\begin{tabular}{l c c c c}
\toprule
\textbf{Model / Method} & \textbf{Prof. Rate} & \textbf{Factuality} & \textbf{Win-Tie Rate} & \textbf{Avg Length} \\
\midrule
Qwen3-235B (+ Business SFT) & 83.20\% & 89.58\% & 50.39\% & 257.58 \\
INS-S1-235B (+ Business SFT) & \textbf{87.07\%} & \textbf{92.47\%} & \textbf{59.27\%} & 261.73 \\
\bottomrule
\end{tabular}
\end{adjustbox}
\end{table}

\noindent\textbf{Analysis:} INS-S1-235B outperforms the business-fine-tuned general baseline across all key metrics. Notably, the \textbf{Human Preference Win-Tie Rate} reached \textbf{59.27\%}, indicating a significant advantage in subjective user experience. Improvements in Professional Rate (+3.87\%) and Factuality Rate (+2.89\%) confirm that initializing from a comprehensive domain paradigm (INS-S1) provides a superior starting point for downstream tasks compared to fine-tuning directly from general models.

\section{Supplementary Figures}

\subsection{Architecture}
This is our detailed architecture diagram.
\label{app:Architecture}
\begin{figure}[htbp]
    \centering
    \includegraphics[width=1\linewidth]{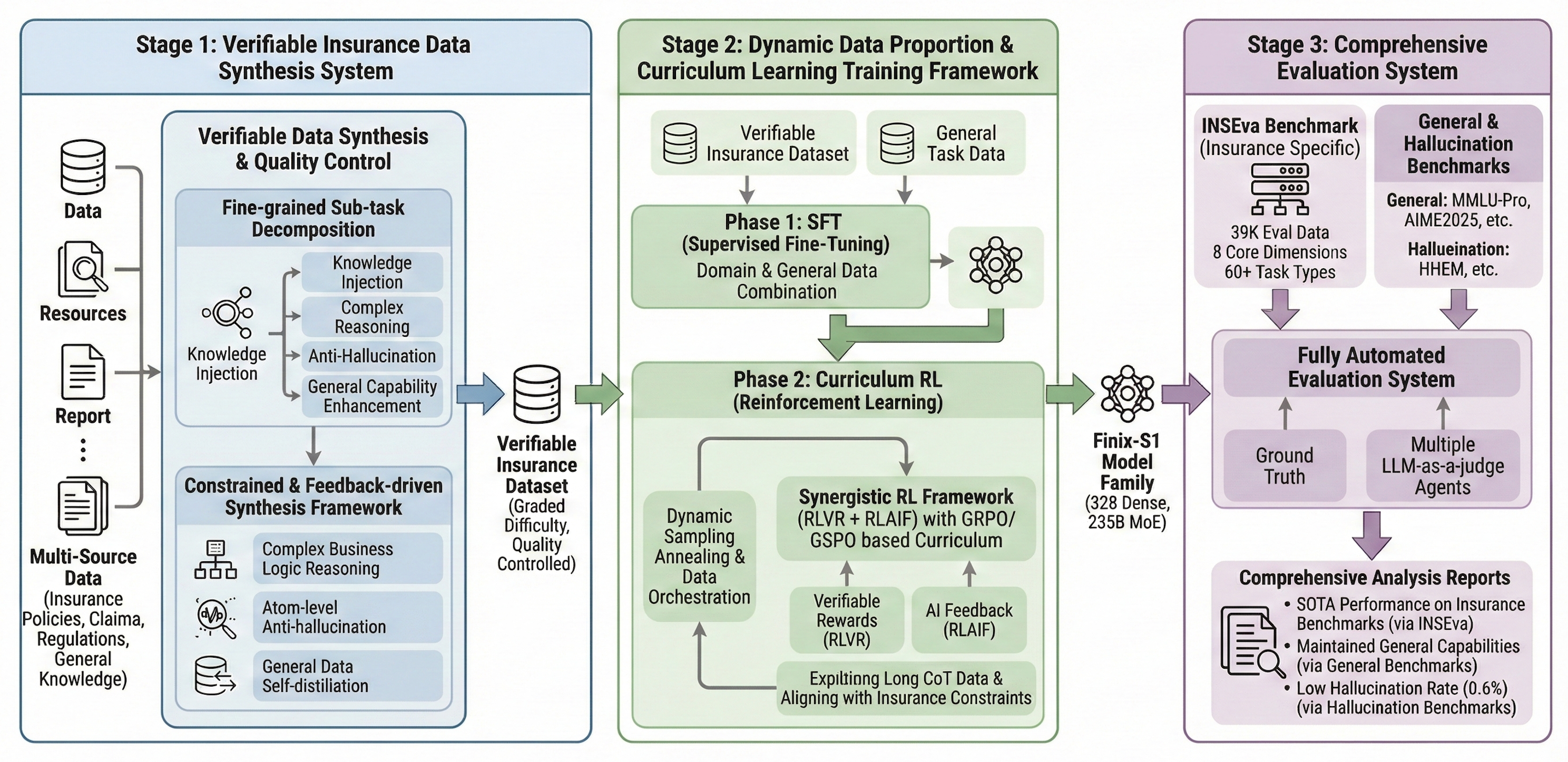}
    \caption{The development process architecture of INS-S1.}
    \label{fig:placeholder}
\end{figure}

\subsection{Data Distribution}
This is a detailed data distribution map of our SFT phase.
\label{app:Data_Distribution}
\begin{figure}[htbp]
    \centering
    \includegraphics[width=0.75\linewidth]{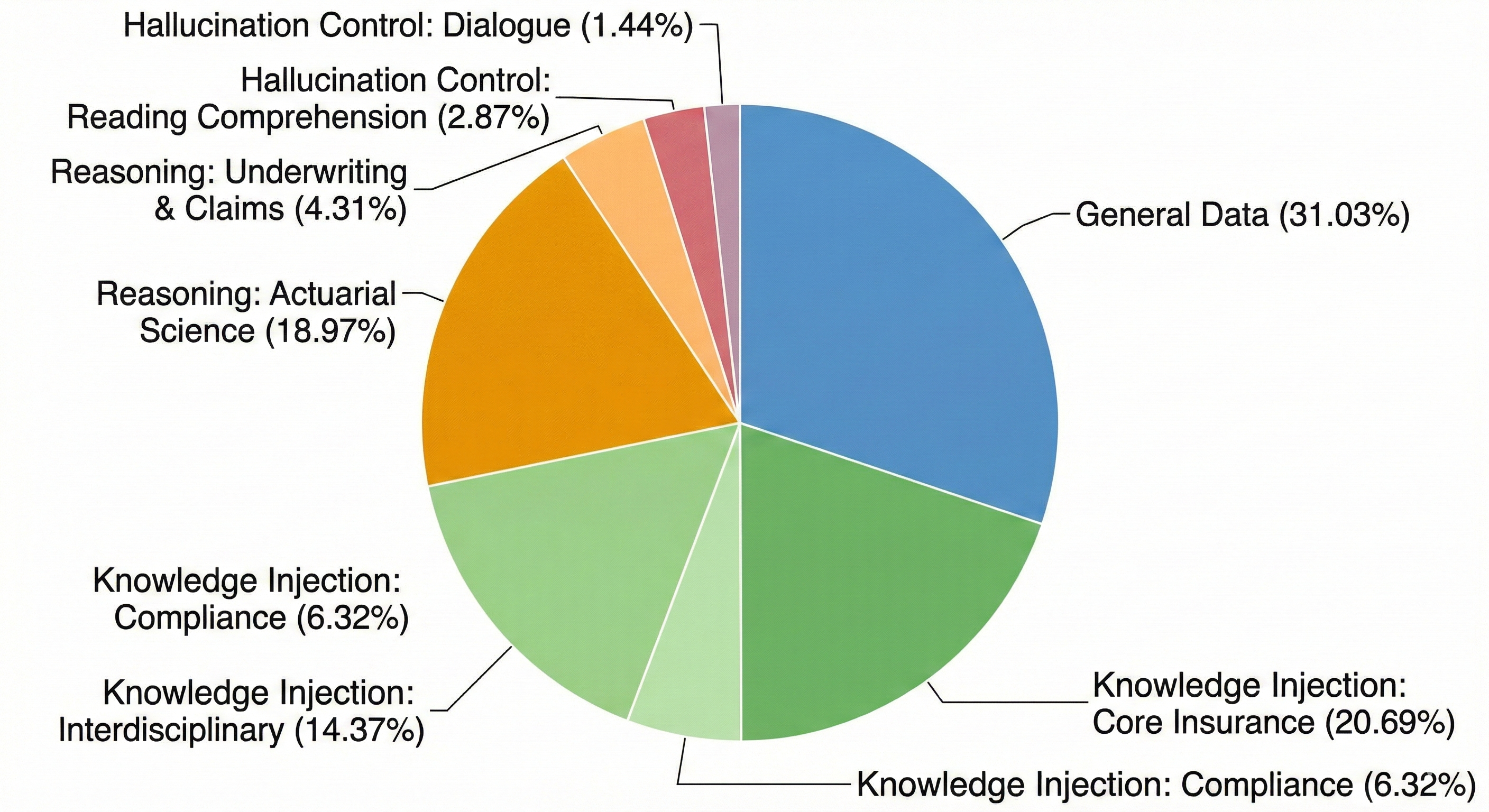}
    \caption{Detailed Data Distribution in SFT Phase}
    \label{fig:placeholder}
\end{figure}

\subsection{Use of RLVR and RLAIF}
This is the specific reward for our RL phase.
\label{app:Use of RLVR and RLAIF}
\begin{figure}[htbp]
    \centering
    \includegraphics[width=0.9\linewidth]{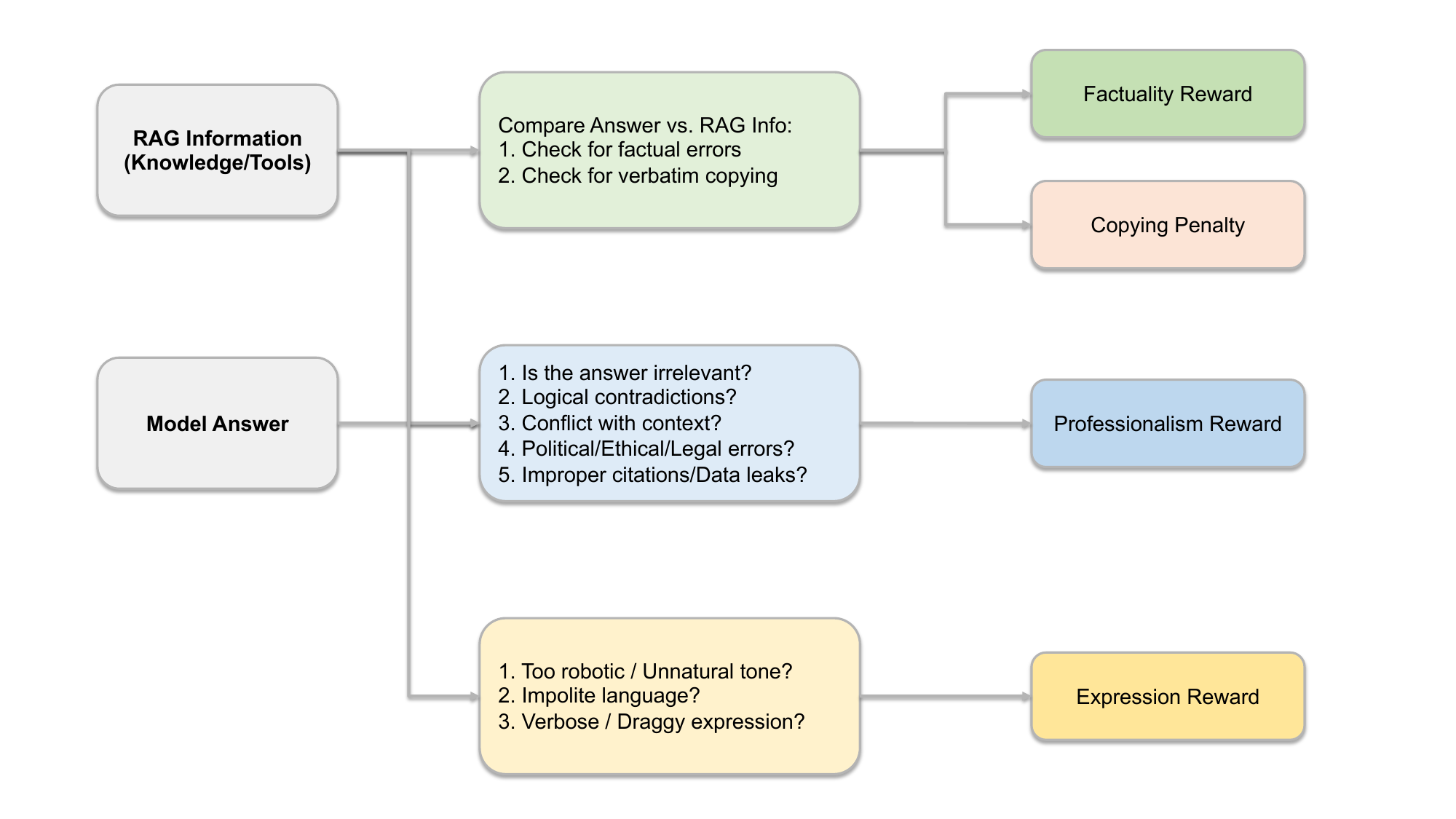}
    \caption{Synergistic Use of RLVR and RLAIF - Reward.}
    \label{fig:placeholder}
\end{figure}

\section{benchmark}

\subsection{INSEva details}
\label{app:INSEva_details}
Our Insurance Business Areas dimensions consists of nine main areas:

\begin{enumerate}
    \item \textbf{Insurance Domain Knowledge (IDK):}
    \begin{itemize}
        \item Insurance Science
    \end{itemize}

    \item \textbf{Insurance-Medicine Interdisciplinary (IMI):}
    \begin{itemize}
        \item Medical Procedure Name Extraction
        \item Medical Condition Name Extraction
        \item Insurance Hospital Name Standardization
        \item Insurance Disease \& Procedure Standardization
        \item Pet Disease Examination Q\&A
        \item Disease Risk Prediction in Medical Reports
    \end{itemize}

    \item \textbf{Insurance Understanding and Cognition (IUC):}
    \begin{itemize}
        \item Insurance Intent Understanding
        \item Insured Object Slot Recognition
        \item Insurance Disease Slot Recognition
        \item Insurance Clause Interpretation
        \item Insurance Product Selection Analysis
        \item Insurance Liability Analysis
        \item Insurance Review Tag Recognition
    \end{itemize}

    \item \textbf{Insurance Logical Reasoning (ILR):}
    \begin{itemize}
        \item Insurance Actuarial Science
        \item Financial Numerical Computation
        \item Insurance Prior Exemption Reasoning
        \item End-to-End Claims Reasoning
        \item Claims Adjudication Decision Reasoning
        \item Underwriting Decision Reasoning
    \end{itemize}

    \item \textbf{Insurance Tool Use (ITU):}
    \begin{itemize}
        \item Insurance Tool Parameter Extraction
        \item Insurance Tool Invocation Timing
    \end{itemize}

    \item \textbf{Insurance Professional Examination (IPE):}
    \begin{itemize}
        \item Insurance Professional Qualification Exam
        \item Practicing Physician Qualification Exam
        \item Practicing Pharmacist Qualification Exam
        \item Insurance Salesperson Certification Exam
        \item Chinese Actuary Examination
    \end{itemize}

    \item \textbf{Insurance Safety and Compliance (ISC):}
    \begin{itemize}
        \item Information Security
        \item Security Baseline
        \item Insurance Document Compliance
        \item Insurance Value System
        \item Insurance Issue Identification
        \item Insurance Fact-Checking
        \item Insurance Compliance Verification
    \end{itemize}

    \item \textbf{Insurance Marketing Growth (IMG):}
    \begin{itemize}
        \item Insurance Service Summary
        \item Insurance Target Audience Positioning
        \item Insurance Marketing Copy Generation
        \item Insurance Configuration Script Generation
    \end{itemize}

    \item \textbf{Insurance Service Dialogues (ISD):}
    \begin{itemize}
        \item \textit{Platform \& Policy Inquiries:}
        \begin{itemize}
            \item Platform-related Inquiries
            \item Regulatory Information Interpretation
            \item Underwriting-related Inquiries
            \item Policy Application-related Inquiries
            \item Premium Payment-related Inquiries
        \end{itemize}
        \item \textit{Claims \& Operations:}
        \begin{itemize}
            \item Claims Assessment-related Inquiries
            \item Claim Settlement-related Inquiries
            \item Post-policy Operations
        \end{itemize}
        \item \textit{Advisory \& Planning:}
        \begin{itemize}
            \item Condition-based Product Selection
            \item Planning and Configuration
            \item Product Recommendation
            \item Insurance Type Comparison
            \item Product Comparison
            \item Benefit \& Yield Calculation
        \end{itemize}
    \end{itemize}
\end{enumerate}

This comprehensive taxonomy covers the full spectrum of insurance business areas, from fundamental domain knowledge to practical service applications, ensuring a thorough evaluation of AI models in the insurance domain.

\subsection{Generation Benchmark Details}
\label{app:Generation_Benchmark_Details}
We evaluate INS-S1 on a comprehensive suite of 40+ benchmarks covering diverse capabilities. 
\textbf{General Knowledge \& Reasoning:} We use MMLU \citep{hendrycks2020measuring}, MMLU-Pro \citep{wang2024mmlupro}, C-Eval \citep{huang2023ceval}, CMMLU \citep{li2023cmmlu}, AGIEval \citep{zhong2023agieval}, and ARC-c \citep{clark2018think}.
\textbf{Mathematics:} Evaluations include GSM8K \citep{cobbe2021training}, AIME24/25 \citep{mathai2025aime25}, and OlympiadBench \citep{he2024olympiadbench}.
\textbf{Coding:} We test on HumanEval \citep{chen2021evaluating}, MBPP+ \citep{liu2023is}, LiveCodeBench \citep{jain2024livecodebench}, and CodeForces \citep{quan2025codeelo}.
\textbf{Instruction Following \& Agent:} We employ IFEval \citep{zhou2023instruction}, BFCL \citep{patil2023gorilla} e.g.

The full results are presented in Table~\ref{tab:full_appendix_results}.

% 这里放你的 sidewaystable 代码

\begin{sidewaystable}[p] % [p] 表示单独一页
    \centering
    \tiny % 使用最小字号
    \setlength{\tabcolsep}{1.2pt} % 极度压缩列间距
    \caption{Full Evaluation Results on General Benchmarks. \textbf{INS-S1} achieves competitive performance compared to significantly larger models.}
    \label{tab:full_appendix_results}

    % 强制将表格宽度缩放到文本高度（因为旋转了，所以是 textheight）
    \resizebox{\textheight}{!}{%
    \begin{tabular}{lccccccccccccccccccccccccccccccccccccccccc}
        \toprule
        \textbf{Model} & \textbf{Avg} & \rotatebox{90}{BFCL\_Live} & \rotatebox{90}{BFCL\_v1} & \rotatebox{90}{Nexus} & \rotatebox{90}{IFEval} & \rotatebox{90}{AlignBench} & \rotatebox{90}{C3} & \rotatebox{90}{WSC} & \rotatebox{90}{Race-H} & \rotatebox{90}{Race-M} & \rotatebox{90}{OCNLI} & \rotatebox{90}{ARC-c} & \rotatebox{90}{Agieval} & \rotatebox{90}{NQ} & \rotatebox{90}{TriviaQA} & \rotatebox{90}{GPQA} & \rotatebox{90}{C-Eval} & \rotatebox{90}{CMMLU} & \rotatebox{90}{MMLU-Pro} & \rotatebox{90}{LogiEval} & \rotatebox{90}{ProntoQA} & \rotatebox{90}{BBH} & \rotatebox{90}{DROP} & \rotatebox{90}{HellaSwag} & \rotatebox{90}{PIQA} & \rotatebox{90}{LCBench} & \rotatebox{90}{LiveCode} & \rotatebox{90}{MBPP+} & \rotatebox{90}{Spider} & \rotatebox{90}{CF} & \rotatebox{90}{Multipl-E} & \rotatebox{90}{CMath} & \rotatebox{90}{MGSM} & \rotatebox{90}{Minerva} & \rotatebox{90}{OlymBench} & \rotatebox{90}{AIME24} & \rotatebox{90}{AIME25} & \rotatebox{90}{LiveBench} & \rotatebox{90}{HLE} & \rotatebox{90}{MT-Bench} & \rotatebox{90}{ArenaHard} \\
        \midrule
        Qwen3-30B-A3B-Inst & 76.40 & 73.13 & 86.89 & 49.37 & 86.81 & 81.23 & 96.22 & 89.42 & 90.42 & 94.50 & 71.36 & 95.93 & 84.26 & 38.78 & 65.67 & 55.73 & 87.87 & 86.61 & 73.58 & 75.02 & 96.62 & 85.37 & 89.21 & 86.17 & 91.84 & 77.26 & 41.41 & 78.84 & 81.60 & 61.15 & 67.62 & 96.45 & 88.00 & 78.72 & 80.69 & 74.58 & 62.29 & 58.89 & 6.07 & 97.40 & 72.94 \\
        Qwen3-235B-A22B-Inst & 79.91 & 76.63 & 89.34 & 56.57 & 88.76 & 84.04 & 97.92 & 96.15 & 93.45 & 95.61 & 73.63 & 95.93 & 88.35 & 41.44 & 76.45 & 59.99 & 91.15 & 90.33 & 78.95 & 75.37 & 96.88 & 89.23 & 90.94 & 91.18 & 95.05 & 83.33 & 48.13 & 80.42 & 85.00 & 66.10 & 70.21 & 97.27 & 93.20 & 81.66 & 75.80 & 78.96 & 68.54 & 60.52 & 9.45 & 97.50 & 86.94 \\
        Kimi-K2-Inst & 78.17 & 72.63 & 81.69 & 48.56 & 90.89 & 81.68 & 97.64 & 96.21 & 92.97 & 95.47 & 69.69 & 95.93 & 87.42 & 49.83 & 82.26 & 72.73 & 90.80 & 89.17 & 81.27 & 75.89 & 95.94 & 89.56 & 92.25 & 90.31 & 95.48 & 80.11 & 47.19 & 79.46 & 83.42 & 55.03 & 57.82 & 96.22 & 91.90 & 81.00 & 71.31 & 62.19 & 48.28 & 64.13 & 5.51 & 96.10 & 90.66 \\
        Ling-1T & 80.70 & 71.16 & 71.95 & 49.03 & 84.83 & 83.27 & 97.53 & 98.26 & 92.74 & 96.17 & 79.39 & 96.02 & 88.45 & 45.48 & 80.13 & 73.77 & 91.77 & 90.53 & 81.02 & 76.92 & 97.94 & 91.85 & 83.49 & 90.56 & 95.38 & 95.36 & 56.00 & 81.38 & 84.16 & 75.89 & 73.68 & 97.24 & 91.70 & 84.68 & 78.94 & 78.18 & 69.69 & 58.55 & 6.30 & 96.70 & 91.82 \\
        GPT-4o & 67.78 & 77.07 & 88.17 & 53.19 & 82.61 & 73.30 & 97.59 & 94.71 & 91.80 & 95.33 & 70.31 & 95.72 & 69.61 & 49.47 & 81.53 & 53.88 & 78.58 & 80.05 & 63.82 & 75.91 & 93.12 & 84.06 & 82.55 & 91.57 & 95.32 & 65.79 & 30.07 & 76.62 & 79.93 & 23.39 & 70.18 & 94.40 & 90.65 & 74.88 & 46.62 & 11.88 & 8.80 & 45.12 & 2.32 & 9.50 & 61.62 \\
        \midrule
        Qwen3-32B (Think) & 79.22 & 76.67 & 89.15 & 45.32 & 86.27 & 78.08 & 96.99 & 96.03 & 91.62 & 94.85 & 76.17 & 95.85 & 84.65 & 35.48 & 68.73 & 66.64 & 88.68 & 87.32 & 78.38 & 77.76 & 96.56 & 88.95 & 86.32 & 85.27 & 92.76 & 90.66 & 60.90 & 82.28 & 81.04 & 80.91 & 66.86 & 97.06 & 92.15 & 81.19 & 82.22 & 80.73 & 71.67 & 60.00 & 8.29 & 97.00 & 71.37 \\
        Qwen3-80B-A3B (Think) & 80.79 & 77.19 & 70.29 & 39.62 & 87.51 & 82.16 & 97.59 & 98.86 & 89.65 & 95.75 & 74.14 & 96.02 & 88.47 & 42.52 & 76.82 & 76.58 & 90.63 & 88.93 & 81.40 & 73.62 & 97.19 & 89.29 & 91.79 & 88.23 & 94.61 & 69.26 & 71.59 & 82.18 & 82.53 & 84.77 & 74.03 & 96.86 & 92.05 & 81.68 & 87.80 & 92.14 & 85.47 & 66.05 & 10.89 & 98.50 & 66.87 \\
        Qwen3-235B-A22B (Think) & 81.40 & 76.53 & 88.75 & 46.43 & 90.89 & 85.49 & 98.30 & 99.52 & 93.05 & 96.03 & 74.00 & 95.55 & 90.27 & 45.12 & 79.51 & 83.30 & 92.46 & 90.59 & 83.30 & 77.34 & 96.25 & 90.78 & 91.61 & 89.39 & 94.18 & 90.23 & 69.77 & 82.41 & 81.88 & 85.62 & 62.75 & 97.45 & 91.90 & 81.86 & 79.76 & 86.20 & 79.64 & 49.59 & 12.33 & 98.40 & 57.70 \\
        DeepSeek-R1 & 82.17 & 76.65 & 87.83 & 39.04 & 82.52 & 85.56 & 98.25 & 99.28 & 92.42 & 95.82 & 77.32 & 96.61 & 87.77 & 48.34 & 80.99 & 80.33 & 90.30 & 88.59 & 83.48 & 78.13 & 97.06 & 88.55 & 90.08 & 86.05 & 95.21 & 93.81 & 71.59 & 81.65 & 76.57 & 91.52 & 76.91 & 96.70 & 91.95 & 76.84 & 77.66 & 88.39 & 84.90 & 49.94 & 13.53 & 97.00 & 91.64 \\
        Ring-1T & 82.34 & 77.31 & 89.10 & 38.68 & 87.05 & 85.24 & 97.37 & 99.22 & 92.71 & 95.61 & 75.02 & 95.59 & 88.38 & 45.26 & 79.07 & 79.17 & 91.20 & 88.98 & 80.47 & 74.90 & 97.19 & 90.68 & 92.02 & 87.82 & 93.96 & 95.71 & 75.33 & 81.32 & 81.84 & 94.14 & 74.96 & 97.04 & 92.35 & 83.58 & 80.38 & 92.60 & 88.65 & 65.18 & 13.95 & 98.20 & 56.51 \\
        Gemini-2.5-Pro & 81.98 & 75.00 & 80.15 & 58.37 & 87.68 & 84.73 & 98.14 & 99.76 & 92.40 & 96.03 & 71.73 & 95.81 & 88.93 & 52.47 & 82.32 & 84.56 & 90.04 & 78.25 & 85.54 & 76.70 & 98.00 & 86.48 & 82.84 & 92.43 & 95.97 & 90.43 & 70.54 & 82.51 & 79.93 & 90.11 & 59.59 & 94.60 & 88.70 & 81.07 & 76.36 & 92.00 & 86.70 & 51.04 & 18.77 & 96.20 & 86.48 \\
        \midrule
        Dianjin-R1-32B & 68.82 & 72.59 & 85.75 & 43.14 & 75.46 & 75.31 & 96.77 & 96.09 & 91.34 & 94.22 & 75.73 & 96.27 & 77.00 & 32.38 & 68.13 & 58.74 & 88.55 & 87.07 & 75.78 & 80.18 & 96.56 & 86.43 & 82.62 & 88.02 & 93.04 & 41.64 & 22.80 & 75.23 & 78.11 & 40.06 & 71.07 & 96.54 & 90.90 & 73.71 & 60.99 & 30.05 & 22.60 & 42.57 & 4.08 & 9.36 & 76.02 \\
        \midrule
        \rowcolor{gray!15} \textbf{INS-S1-32B} & 80.08 & 75.77 & 88.47 & 65.86 & 85.38 & 78.56 & 96.49 & 93.27 & 89.99 & 94.43 & 74.37 & 95.68 & 86.09 & 33.74 & 68.27 & 63.04 & 88.87 & 86.50 & 77.73 & 85.82 & 97.12 & 88.97 & 93.13 & 85.66 & 93.31 & 90.97 & 62.83 & 81.58 & 81.93 & 77.62 & 77.39 & 96.47 & 92.30 & 81.25 & 82.12 & 84.84 & 76.41 & 61.22 & 6.38 & 96.70 & 66.73 \\
        \rowcolor{gray!15} \textbf{INS-S1-235B} & 81.77 & 77.14 & 89.24 & 54.00 & 86.69 & 83.89 & 96.77 & 98.62 & 93.60 & 96.17 & 66.78 & 96.19 & 87.76 & 45.32 & 78.42 & 81.22 & 91.53 & 88.50 & 82.26 & 78.33 & 95.44 & 92.17 & 91.35 & 88.45 & 94.12 & 93.95 & 71.81 & 82.44 & 84.16 & 85.17 & 65.53 & 97.15 & 92.00 & 84.13 & 86.42 & 87.76 & 80.21 & 52.23 & 12.79 & 98.30 & 62.80 \\
        \bottomrule
    \end{tabular}
    }
\end{sidewaystable}

\end{document}

%% file: math_commands.tex
\usepackage{amsmath,amsfonts,bm}

\def\eqref#1{equation~\ref{#1}}
\def\1{\bm{1}}

\DeclareMathAlphabet{\mathsfit}{\encodingdefault}{\sfdefault}{m}{sl}
\SetMathAlphabet{\mathsfit}{bold}{\encodingdefault}{\sfdefault}{bx}{n}